\documentclass{egpubl}
\usepackage{egsr2024}
 
\SpecialIssuePaper         

\CGFStandardLicense

\usepackage[T1]{fontenc}
\usepackage{dfadobe}  
\usepackage{booktabs} 
\usepackage{enumitem}
\usepackage{microtype}
\usepackage{colortbl}
\usepackage[table]{xcolor}

\usepackage{amssymb}
\usepackage{amsbsy}
\usepackage{amsmath}

\usepackage[export]{adjustbox}
\usepackage[ruled]{algorithm2e} 
\usepackage[normalem]{ulem} 
\usepackage{hyphenat}

\SetAlFnt{\small}
\SetAlCapFnt{\small}
\SetAlCapNameFnt{\small}
\SetAlCapHSkip{0pt}

\newcommand{\NEW}[1]{#1}

\newcommand{\REALTIME}{interactive } 
\newcommand{\INREALTIME}{interactively } 

\usepackage{cite}  
\BibtexOrBiblatex
\electronicVersion
\PrintedOrElectronic

\usepackage{egweblnk} 

\begin{document}

\title[A Diffusion Approach to Radiance Field Relighting]{A Diffusion Approach to Radiance Field Relighting using Multi-Illumination Synthesis}


%

\teaser{
    \vspace{-2.5cm}
    \author[Y. Poirier-Ginter et al.]{}
    {\parbox{\textwidth}{\centering Y. Poirier-Ginter$^{1,2}$\orcid{0009-0001-5806-5136}, A. Gauthier$^{1}$\orcid{0000-0002-3710-0879}, J. Philip$^{3}$\orcid{0000-0003-3125-1614}, J.-F. Lalonde$^{2}$\orcid{0000-0002-6583-2364} and G. Drettakis$^{1}$\orcid{0000-0002-9254-4819}}}
    {\parbox{\textwidth}{\centering 
    $^1$Inria, Université Côte d'Azur, France; $^2$Université Laval, Canada; $^3$Adobe Research, United Kingdom\\}}
    \vspace{0.5cm}
    \centering
 
    \includegraphics[width=0.95\linewidth]{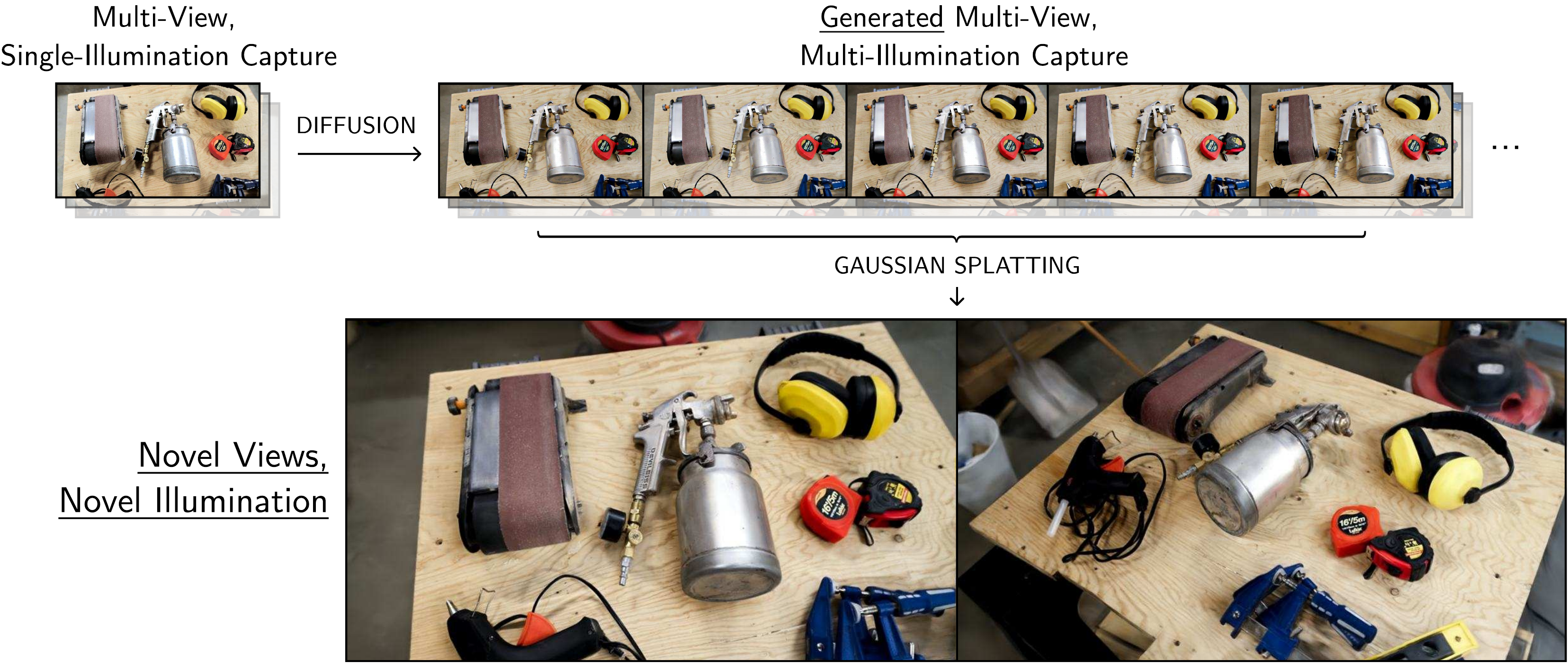}
    \caption{
    Our method produces relightable radiance fields directly from single-illumination multi-view dataset, by using priors from generative data in the place of an actual multi-illumination capture.}
\label{fig:teaser}
}

\maketitle



%

\begin{abstract}
Relighting radiance fields is severely underconstrained for multi-view data, which is most often captured under a single illumination condition; It is especially hard for full scenes containing multiple objects. We introduce a method to create relightable radiance fields using such single-illumination data by exploiting priors extracted from 2D image diffusion models. We first fine-tune a 2D diffusion model on a multi-illumination dataset conditioned by light direction, allowing us to augment a single-illumination capture into a realistic -- but possibly inconsistent -- multi-illumination dataset from directly defined light directions.
We use this augmented data to create a relightable radiance field represented by 3D Gaussian splats. To allow direct control of light direction for low-frequency lighting, we represent appearance with a multi-layer perceptron parameterized on light direction. To enforce multi-view consistency and overcome inaccuracies we optimize a per-image auxiliary feature vector. We show results on synthetic and real multi-view data under single illumination, demonstrating that our method successfully exploits 2D diffusion model priors to allow realistic 3D relighting for complete scenes.


\end{abstract}

\section{Introduction}

Radiance fields have recently revolutionized 3D scene capture from images~\cite{nerf}.
Such captures typically involve a multi-view set of photographs taken under the \emph{same} lighting conditions.
\emph{Relighting} such radiance fields is hard since lighting and material properties are entangled (e.g., is this a shadow or simply a darker color?) and the inverse problem ill-posed. 

One approach to overcome this difficulty is to capture a multi-illumination dataset which better conditions the inverse problem but comes at the cost of a heavy capture setup~\cite{debevec2000acquiring}.
Another option is to use \emph{priors}, which is typically done by training a neural network on synthetic data to predict intrinsic properties or relit images. 
However, creating sufficiently large, varied and photorealistic 3D scenes is both challenging and time-consuming.
As such, methods relying on these---or simpler---priors often demonstrate results on isolated masked objects~\cite{boss2021nerd}, or make simplifying assumptions such as distant environment lighting~\cite{boss2021neural,zhang2021nerfactor}.
Other methods have handled more complex illumination models, including full scenes~\cite{philip2021free,philip2019multi}, but can be limited in the complexity of the geometry and materials that must reconstruct well.
Finally, methods that depend on accurate estimates of surface normals~\cite{jin2023tensoir,R3DG2023} often produce limited levels of realism when relighting.

At the other end of the spectrum, diffusion models (DMs, e.g., \cite{stable-diffusion}), trained on billions of natural images, have shown exceptional abilities to capture real image distribution priors and can synthesize complex lighting effects. While recent progress shows they can be controlled in various ways~\cite{controlnet}, extracting lighting-specific priors from these models, especially for full 3D scenes, has not yet been demonstrated. 


In this paper, we build on these observations and present a new method that demonstrates that it is possible to create relightable radiance fields for complete scenes from single low-frequency lighting condition captures by exploiting 2D diffusion model priors.
We first propose to fine-tune a pre-trained DM conditioned on the dominant light source direction. 
For this, we leverage a dataset of images with many lighting conditions of the same scene~\cite{multilum}, which enables the DM to produce relit versions of an image with explicit control over the dominant lighting direction.
We use this 2D relighting network to augment any standard multi-view dataset taken under single lighting by generating multiple relit versions of each image, effectively transforming it into a multi-illumination dataset.
Given this augmented dataset, we train a new \emph{relightable radiance field} with direct control on lighting direction, which in turn enables realistic \REALTIME relighting of full scenes with lighting and camera view control in real time for low-frequency lighting. 
We build on 3D Gaussian Splatting~\cite{gaussian-splats}, enhancing the radiance field with a small Multi-Layer Perceptron and an auxiliary feature vector to account for the approximate nature of the generated lightings and to handle lighting inconsistencies between views.

\noindent
In summary, our contributions are:
\begin{itemize}[nosep]
	\item A new 2D relighting neural network with direct control on lighting direction, created by fine-tuning a DM with multi-lighting data.
	\item A method to augment single-lighting multi-view capture to an approximate multi-lighting dataset, by exploiting the 2D relighting network.
	\item An \REALTIME relightable radiance field that provides direct control on lighting direction, and corrects for inconsistencies in the neural relighting.
\end{itemize}
\noindent
We demonstrate our solution on synthetic and real indoor scenes, showing that it provides realistic relighting of multi-view datasets captured under a single lighting condition in real time.

\section{Related Work}

Our method proposes a relightable radiance field. We review work on radiance fields and their relightable variants, and discuss diffusion models and fine-tuning methods we build on.

\subsection{Radiance Fields}

Radiance field methods have revolutionized 3D scene capture using multi-view datasets (photos or video) as input. 
In particular, Neural Radiance Fields (NeRFs)~\cite{nerf} learn to synthesize novel views of a given scene by regressing its radiance from a set of input images (multiple photos or videos of a 3D scene).
Structure from motion~\cite{sfm,colmap1} is used to estimate the camera poses for all images and rays are cast through the center of all pixels.
A multi-layer perceptron (MLP) $\boldsymbol{c}_{\theta}$ parameterized by 3D position and view direction is used to represent the radiance and opacity of the scene.
The optimization objective is simply the mean squared error:
$$
\mathcal{L}_{\mathrm{NeRF}} = 
\mathbb{E}_{\boldsymbol{o}, \boldsymbol{d}, \boldsymbol{c}^\ast}\Big[||\boldsymbol{c}_{\theta}(\boldsymbol{o}, \boldsymbol{d}) - \boldsymbol{c}^\ast||_2^2 \Big] \,,
$$
where $\boldsymbol{o}$ is a ray's origin, $\boldsymbol{d}$ its direction, and $\boldsymbol{c}^\ast$ the target RGB color value of its corresponding pixel.
The predicted color for that pixel is obtained by integrating a color field $\boldsymbol{c}_{\theta}$ weighted by a density field $\sigma_{\theta}$ following the equation of volume rendering. 
The original NeRF was slow to train and to render; A vast number of methods~\cite{tewari2022advances} have been proposed to improve the original technique, e.g., acceleration structures~\cite{mueller2022instant}, antialiasing~\cite{barron2021mipnerf}, handling larger scenes~\cite{mipnerf360} etc.
Recently, 3D Gaussian Splatting (3DGS)~\cite{gaussian-splats} introduces an explicit, primitive-based representation of radiance fields.
The anisotropic nature of the 3D Gaussians allows the efficient representation of fine detail, and the fast GPU-accelerated rasterization used allows real-time rendering.
We use 3DGS to represent radiance fields mainly for performance, but any other radiance representation, e.g., \cite{chen2022tensorf}, could be used instead.
Radiance fields are most commonly used in the context of \emph{single-light condition} captures, i.e., the images are all captured under the same lighting.
As a result, there is no direct way to change the lighting of captured scenes, severely restricting the utility of radiance fields compared to traditional 3D graphics assets.
Our method uses diffusion models to simulate multi-light conditions from a single-light capture thus allowing the relighting of radiance fields.
%

\subsection{Single Image Relighting}

Single image relighting approaches have mostly been restricted to human faces, with the most recent methods using generative priors \cite{wang2008face,shu2017neural,sengupta2018sfsnet,futschik2023controllable,ponglertnapakorn2023difareli,Papantoniou2023RelightifyR3}. Recently, human body relighting has also been studied ~\cite{kanamori2019relighting,lagunas2021single} as their structure also allows for the design of effective priors.
Because they are much less constrained, relighting generic scenes from a single image is a much harder problem that has eluded researchers until recently.
While some approaches have focused on specific relighting effects such as shadows~\cite{liu2020arshadowgan,sheng2022controllable,sheng2021ssn,valenca2023shadow}, they are applicable solely for the purpose of object compositing.
Full scene relighting has been explored by Murmann et al.~\shortcite{multilum}, who present a dataset of real indoor scenes lit by multiple lighting directions.
They show that training a U-net on their dataset allows for full scene relighting---in this work, we also leverage their dataset but train a more powerful ControlNet~\cite{controlnet} for the relighting task.
Other works include \cite{ture2021noon} who focus on sky relighting and \cite{li2022physically,zhu2022designing} which leverage image-to-image translation.
Methods for outdoor scenes have also been proposed~\cite{yu2020self,liu2020learning,yu2022outdoor}.
Of note, SIMBAR~\cite{simbar22} 
 and OutCast~\cite{griffiths2022outcast} produces realistic, user-controllable, hard, cast shadows from the sun.
In contrast, we focus on indoor scenes which often exhibit soft shadows and more complex lighting effects. \NEW{Finally, the concurrent work of Zeng et al.~\cite{zeng2024dilightnet} uses diffusion models to relight isolated objects using environment maps. In contrast to these solutions, we focus on cluttered indoor scenes which often exhibit soft shadows and more complex lighting effects.}

\subsection{Multi-view Relighting}
While single-view methods produce good results on restricted datasets such as faces, they are often limited by the lack of accurate geometry, required to simulate light transport.
To this point, multi-view data can provide a more accurate and complete geometric reconstruction.
For example Philip et al.~\shortcite{philip2021free,philip2019multi} build on multi-view stereo (MVS) reconstruction of the scene, and learn a prior from synthetic data rendered under multiple lighting conditions.
Despite correcting for many of the reconstruction artifacts, these methods are restricted by the quality of MVS reconstruction.
Nimier et al. \cite{nimier2021material} also present a scene-scale solution but require a complex pipeline that optimizes in texture space. Gao et al.~\cite{deferredNeural20} use a rough proxy and neural textures to allow object relighting.

More recently radiance fields have also been used as a geometric representation for relighting.
Most methods work on the simple case of a single isolated object while we target larger scenes.
Such methods typically assume lighting to be distant, often provided as an environment map.
NeRFactor~\cite{zhang2021nerfactor} uses a Bi-Directional Reflectance Distribution Function (BRDF) prior from measurements and estimates normals and visibility, while NERV~\cite{nerv2021} and Zhang et al.~\cite{zhang2022invrender} predicts visibility to allow indirect illumination estimation.
NERD~\cite{boss2021nerd}, PhySG~\cite{physg2021}, and DE-NeRF~\cite{DE-NeRF} use efficient physically-based material and lighting models to decompose a radiance field into spatially varying BRDFs, while Neural-PIL~\cite{boss2021neural} learns the illumination integration and low-dimensional BRDF priors using auto-encoders.
TensorIR~\cite{jin2023tensoir} uses a mixed radiance and physics-based formulation to recover intrinsic properties.
NeRO~\cite{liu2023nero} focuses on specular objects showing very promising results.
Relightable Gaussians~\cite{R3DG2023} use the more recent 3D Gaussian representation along with ray-tracing to estimate properties of objects. 
GS-IR \cite{gs-ir}, GaussianShader~\cite{gaussianshader} and GIR~\cite{shi2023gir} also build on 3D Gaussian splatting, proposing different approaches to estimate more reliable normals while approximating visibility and indirect illumination; these work well for isolated objects under distant lighting. However, these methods struggle with more complex scene-scale input and near-field illumination but can work or be adapted to both single and multi-illumination input data.

Feeding multi-view multi-illumination data to a relightable radiance field indeed enables better relighting but at the cost of highly-controlled capture conditions  \cite{xu2023renerf, shadowHints23, Rene_2023_CVPR} or an extended dataset of unconstrained illuminations \cite{boss2022-samurai,li2022neulighting}.
In our method, we use a Diffusion Model to simulate multi-illumination data, lifting the capture constraints while benefiting from the lighting variations.
Another body of work 
\cite{Munkberg_2022_CVPR, hasselgren2022shape, yao2022neilf, zhang2023neilfpp, li2022texir, Liang_2023_ICCV} achieve object or scene relighting from multi-view images by extracting traditional 3D assets (meshes and SVBRDFs) and applying physically-based rendering algorithms. 
IBL-NeRF \cite{iblnerf2023} allows for scene-scale material estimation but bakes the illumination into a prefiltered light-field which prevents relighting.
Recently, NeRF-OSR \cite{rudnev2022nerf}, $\text{I}^\text{2}$-SDF \cite{zhu2023i2}, and Wang et al.~\cite{wang2023fegr} focused on scene scale, single illumination relighting scenes using both implicit and explicit representations.
While they can achieve reasonable results, they often lack overall realism, exhibiting bumpy or overly smooth shading during relighting.
In contrast, our use of diffusion priors provides realistic-looking output.

\begin{figure*}[!h]
    \includegraphics[width=\linewidth]{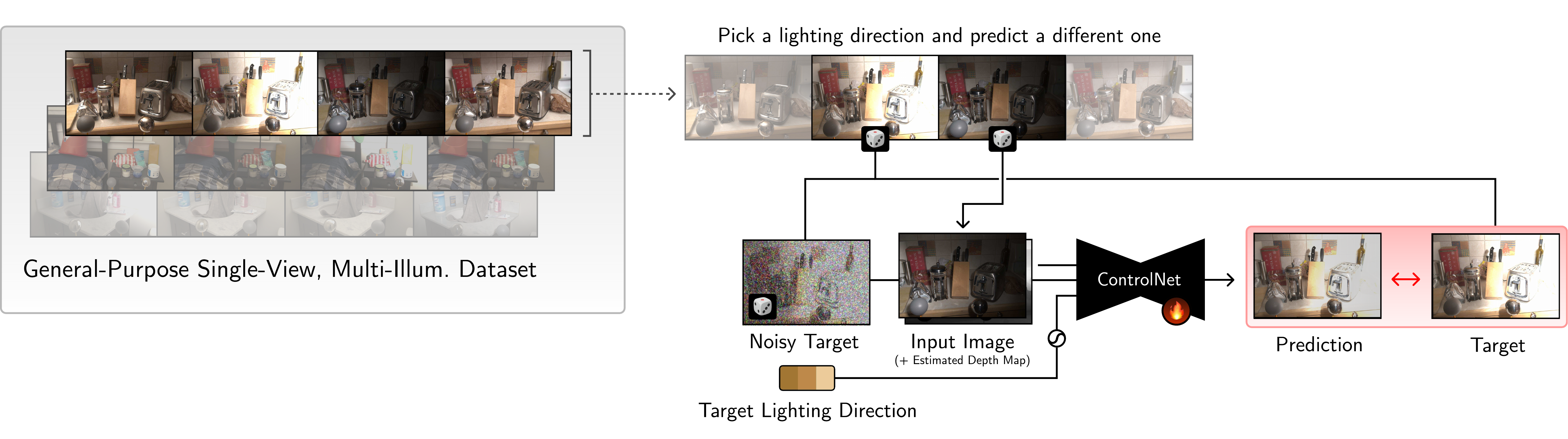}
    \caption{We use the single-view, multi-illumination dataset of Murmann et al.~\shortcite{multilum} to train ControlNet~\cite{controlnet} on single view supervised relighting. The network accepts an image (along with its estimated depth map) and a target light direction as input and produces a relit version of the same scene under the desired target lighting.}
    \label{fig:controlnet-relighting}
    
\end{figure*}

\subsection{Diffusion Models}

Diffusion Models (DMs)~\cite{thermo, ddpm} made it possible to train generative models on diverse, high-resolution datasets of billions of images.
These models learn to invert a forward diffusion process that gradually transforms images into isotropic Gaussian noise, by adding random Gaussian noise $\boldsymbol{\epsilon}_t \sim \mathcal{N}(\boldsymbol{0}, \boldsymbol{I})$ to an image in $T$ steps.
%
DMs train a neural network $\boldsymbol{g}_\phi$ with parameters $\phi$ to learn to denoise with the objective:
$$
\mathcal{L}_{\mathrm{Diffusion}} = \mathbb{E}_{\boldsymbol{x}, \boldsymbol{\epsilon}, 1 \leq t \leq T}\Big[ ||\boldsymbol{g}_\phi(\boldsymbol{x}_t|t) - \boldsymbol{y}_t||_2^2 \Big] \,,
$$
in which target $\boldsymbol{y}_t$ is often set to $\boldsymbol{\epsilon}$. 
%
After training, sampling can be performed step-by-step, by predicting $\boldsymbol{x}_{t-1}$ from $\boldsymbol{x}_t$ for each timestep $t$ which is expensive since $T$ can be high (e.g., 1000); faster alternatives include deterministic DDIM~\cite{ddim} sampling, that can perform sampling of comparable quality with fewer steps (i.e., $10$-$50\times$ larger steps).
%
Stable Diffusion~\cite{stable-diffusion} performs denoising in a lower-dimensional latent space, by first training a variational encoder to compress images; for instance, in Stable Diffusion XL~\cite{sdxl}, images are mapped to a latent space of size $\mathbb{R}^{128 \times 128 \times 4}$.
In a pre-pass, the dataset is compressed using this autoencoder, and a text-conditioned diffusion model is then trained directly in this latent space. 

Diffusion models have an impressive capacity to synthesize highly realistic images, typically conditioned on text prompts. The power of DMs lies in the fact that the billions of images used for training contain an extremely rich representation of the visual world.
However, \emph{extracting} the required information for specific tasks, without incurring the (unrealistic) cost of retraining DMs is not straightforward.
A set of recent methods show that it is possible to fine-tune DMs with a typically much shorter training process to perform specific tasks (e.g.,~\cite{gal2023encoder,ruiz2023dreambooth}).
A notable example is ControlNet~\cite{controlnet} which proposed an efficient method for fine-tuning Stable Diffusion with added conditioning.
In particular, they demonstrated conditional generation from depth, Canny edges, etc., with and without text prompts; We will build on this solution for our 2D relighting method.

In a similar spirit, there has been significant evidence in recent years that latent spaces of generative models encode material information~\cite{bhattad2023stylegan,bhattad2022stylitgan}.
Recent work shows the potential to fine-tune DMs to allow direct material editing ~\cite{sharma2023alchemist}.
Nonetheless, we are unaware of published methods that use DM fine-tuning to perform realistic relighting of full and cluttered scenes.




\section{Method}

Our method is composed of three main parts. First, we create a 2D relighting neural network with direct control of lighting direction (Sec.~\ref{sec:2d-relight}).
Second, we use this network to augment a multi-view capture with single lighting into a multi-lighting dataset, by using our relighting network.
The resulting dataset can be used to create a radiance field representation of the 3D scene (Sec.~\ref{sec:augmenting}).
Finally, we create a relightable radiance field that accounts for inaccuracies in the synthesized relit input images and provides a multi-view consistent lighting solution (Sec.~\ref{sec:correcting}). 

\subsection{Single-View Relighting with 2D Diffusion Priors}
\label{sec:2d-relight}



Relighting a scene captured under a single lighting condition is severely underconstrained, given the lighting/material ambiguity, and thus requires priors about how appearance changes with illumination.
Arguably, large DMs must internally encode such priors since they can generate realistic complex lighting effects, but existing architectures do not allow for explicit control over lighting. 

We propose to provide explicit control over lighting by fine-tuning a pre-trained Stable Diffusion (SD)~\cite{stable-diffusion} model using ControlNet~\cite{controlnet} on a multi-illumination dataset.
As illustrated in Fig.~\ref{fig:controlnet-relighting}, the ControlNet accepts as input an image as well as a target light direction, and produces a relit version of the same scene under the desired lighting.
To train the ControlNet, we leverage the dataset of Murmann et al.~\shortcite{multilum}, which contains $N=1015$ real indoor scenes captured from a single viewpoint, each lit under $M=25$ different, controlled lighting directions. We only keep the 18 non-front facing light directions.

\subsubsection{Lighting Direction}

\begin{figure}[h!]
    \includegraphics[width=\linewidth]{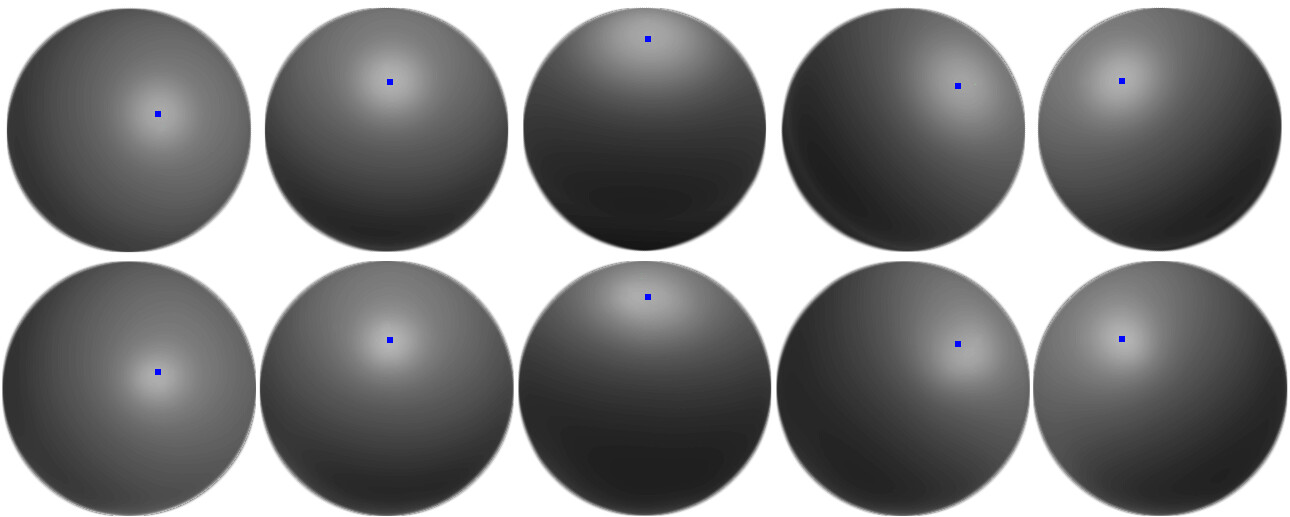}
    \caption{Top row: five diffuse sphere rendered by our optimized lighting direction and shading parameters --- the direction is indicated by a blue dot at the point of maximum specular intensity; Bottom row: the corresponding target gray spheres obtained by averaging the diffuse spheres captured in all spheres. We found the lighting directions by minimizing the $L_1$ distance between the top and bottom row.}
    \label{fig:recover-directions}
\end{figure}

To capture the scenes using similar light directions, Murmann et al. relied on a camera-mounted directional flash controlled by a servo motor. 
A pair of diffuse and metallic spheres are also visible in each scene; we leverage the former to obtain the effective lighting directions. Using as target the average of all diffuse spheres produced by the same flash direction, we find the lighting direction $l \in \mathbb{R}^3$ which best reproduces this target when rendering a gray ball with a simplistic Phong shading model. More specifically, we minimize the $L_1$ error when jointly optimizing for an ambient light term and shading parameters (albedo, specular intensity and hardness, as well as a Fresnel coefficient). Fig.~\ref{fig:recover-directions} illustrates this process. 

\subsubsection{Controlling Relighting Diffusion}
We train ControlNet to predict relit versions of the input image by conditioning it on a target lighting direction. Let us denote a set $\mathcal{X}$ of images of a given scene in the multi-light dataset of Murmann et al.~\cite{multilum}, where each image $\mathbf{X}_k \in \mathcal{X}$ has associated light direction $l_k$. 
Our approach, illustrated in Fig.~\ref{fig:controlnet-relighting}, trains on pairs of lighting directions of the same scene (including the identity pair). 
The denoising objective becomes 
\begin{equation}
\mathcal{L}_{\mathrm{2D}} = \mathbb{E}_{\boldsymbol{\epsilon}, \mathbf{X}, t, i, j}\Big[||\boldsymbol{g}_{\psi}(\mathbf{X}_{t,i}; t, \mathbf{X}_j, \mathbf{D}_j, \boldsymbol{l}_i) - \boldsymbol{y}_t||^2_2 \Big] \,,
\end{equation}
where $\mathbf{X}_{t,i}$ is the noisy image at timestep $t \in [1,T]$, where $i, j \in [1,M]$, and where $\psi$ are the ControlNet optimizable parameters only. $\mathbf{X}_j$ is another image from the set and $\mathbf{D}_j$ is its depth map (obtained with the approach of Ke et al.~\cite{marigold})---both are given as input to the ControlNet subnetwork. In short, the network is trained to denoise input image $\mathbf{X}_i$ given its light direction $l_i$ while conditioned on the image $\mathbf{X}_j$ corresponding to another lighting direction $l_j$ of the same scene. Here, we do not use text conditioning: the empty text string is provided to the network. 



%
Specifically, the light direction $\boldsymbol{l}_i$ is encoded using the first 4 bands of spherical harmonics, following the method of M\"uller et al.~\cite{mueller2022instant}.
The resulting vector is added to the timestep embedding prior to feeding it to the layers of ControlNet's trainable copy.


\subsubsection{Improving the Diffusion Quality}
\begin{figure}[h!]
    \includegraphics[width=0.99\linewidth]{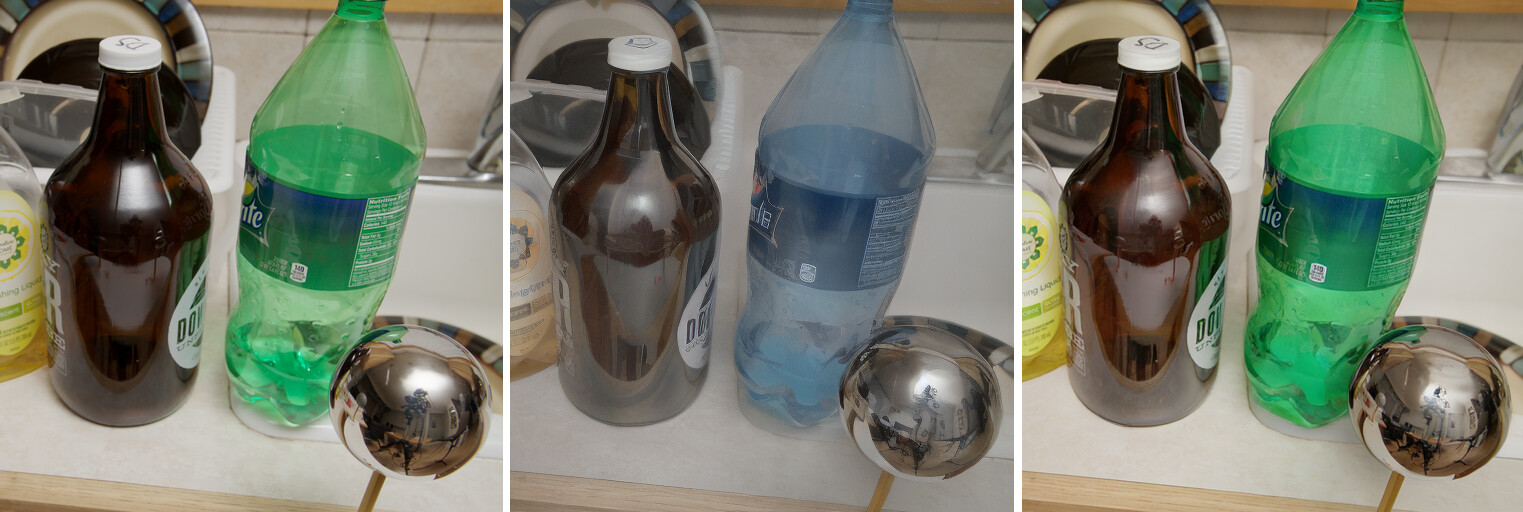}
    \caption{Importance of post-relighting color and contrast adjustments. Left: input image. Middle: naive ControlNet relighting; the bottle has the wrong color and the contrast is poor. Right: our relighting after training with \cite{flawed} and after color-matching the input.}
    \label{fig:improve-color}
\end{figure}

\begin{figure}[h!]
    \includegraphics[width=\linewidth]{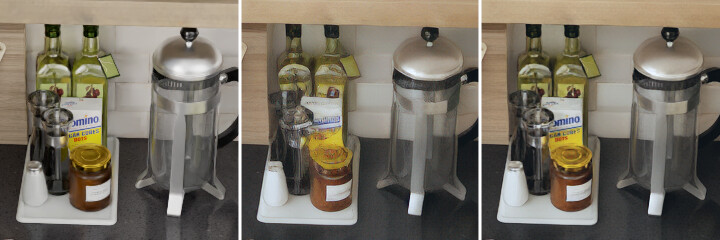}
    \caption{Importance of conserving edge sharpness when relighting. Left: input image. Middle: naive ControlNet relighting; note how the edges do not match the input and how the text is illegible. Right: our final relighting after fine-tuning the conditonal decoder network from~\cite{zhu2023designing}.}
    \label{fig:improve-edges}
\end{figure}

Since ControlNet was not specifically designed for relighting, adapting it naively as described above leads to inaccurate colors and a loss in contrast (see Fig.~\ref{fig:improve-color}), as well as distorted edges (see Fig.~\ref{fig:improve-edges}). These errors also degrade multi-view consistency.

We adopt two strategies to improve coloration and contrast. First, we follow the recommendations of~\cite{flawed} to improve image brightness---we found them to also help for color. In particular, using the ``v-parameterized'' objective $y_t = \sqrt{\bar\alpha_t} \cdot \boldsymbol{\epsilon} - \sqrt{1 - \bar\alpha_t} \cdot \boldsymbol{x}$, instead of the more usual $y_t = \epsilon$, proved critical; in this equation, $1 - \bar\alpha_t$ gives the variance of the noise at timestep $t$. Second, after sampling, we color-match predictions to the input image to compensate for the difference between the color distribution of the training data and that of the scene. This is done by subtracting the per-channel mean and dividing by the standard deviation for the prediction, then adding the mean and standard deviation of the input, in the LAB colorspace. This is computed over all 18 lighting conditions together (i.e., the mean over all lighting directions) to conserve relative brightness across all conditions. Fig.~\ref{fig:improve-color} shows the effect of these changes; without them, the bottle is blue instead of green and overall contrast is poor.

To correct the distorted edges, we adapt the asymmetric autoencoder approach of Zhu et al.~\cite{zhu2023designing}, which consists in conditioning the latent space decoder with the (masked) input image for the inpainting task. In our case, we ignore the masking and fine-tune the decoder on the multi-illumination dataset \cite{multilum}. At each fine-tuning step, we encode an image and condition the decoder on an image from the same scene with another random lighting direction. The decoder is fine-tuned with the Adam optimizer at a learning rate of $10^{-4}$ for 20k steps when training at resolution $768 \times 512$ and 50k steps at resolution $1536 \times 1024$. Note that this step is independent of the ControlNet training. Fig.~\ref{fig:improve-edges} shows the effect of these changes; note how the edges are wobbly and the text is illegible without them.

\begin{figure*}[!th]
	\footnotesize
	\setlength{\tabcolsep}{0.4pt}
    \begin{tabular}{cccc}
    Input Image & 
    \multicolumn{3}{c}{ControlNet Relightings} \\
    
    \includegraphics[width=0.248\linewidth]{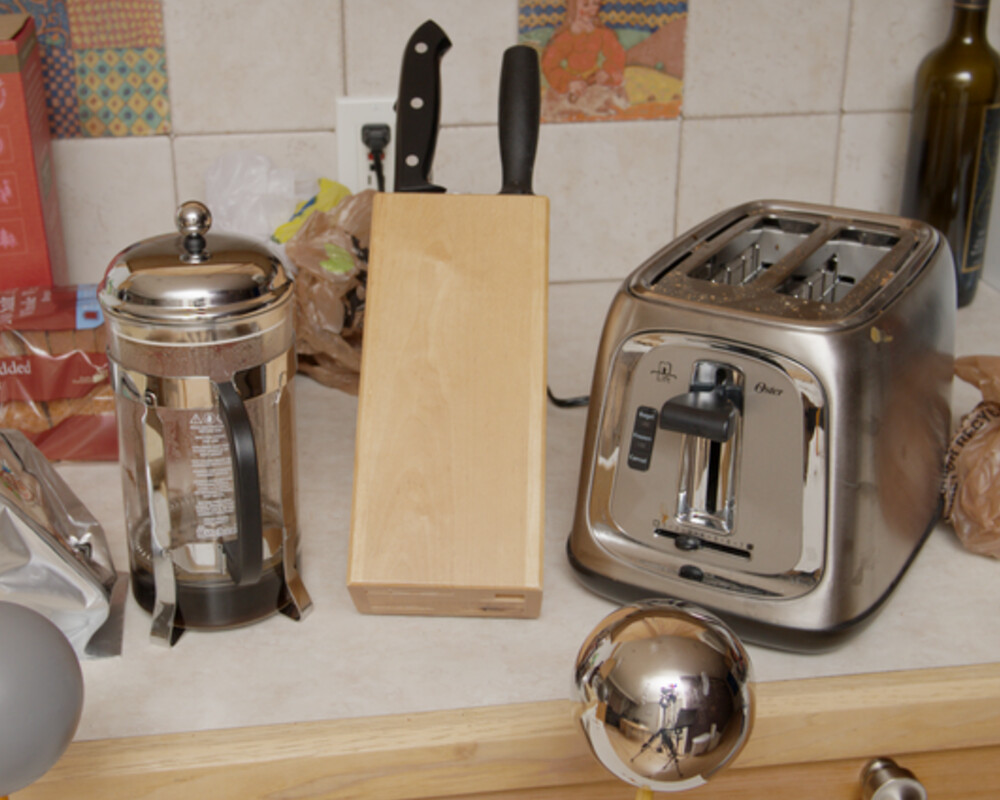} & 
    \setlength{\tabcolsep}{0.25pt}
    \includegraphics[width=0.248\linewidth]{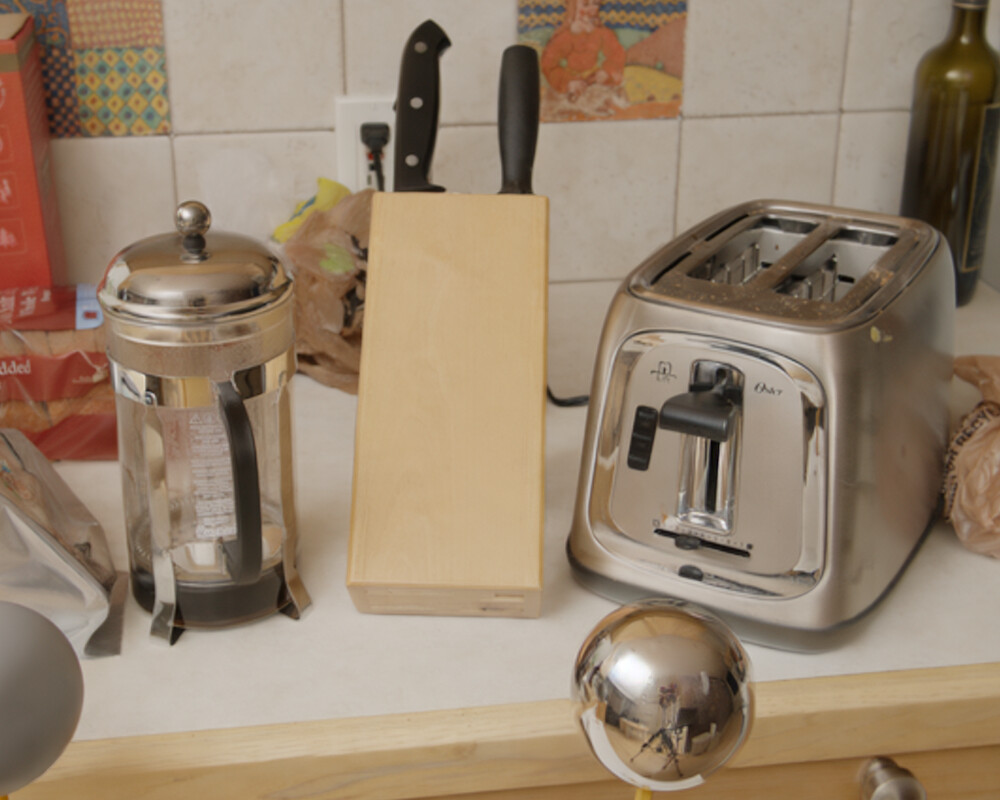} & 
    \includegraphics[width=0.248\linewidth]{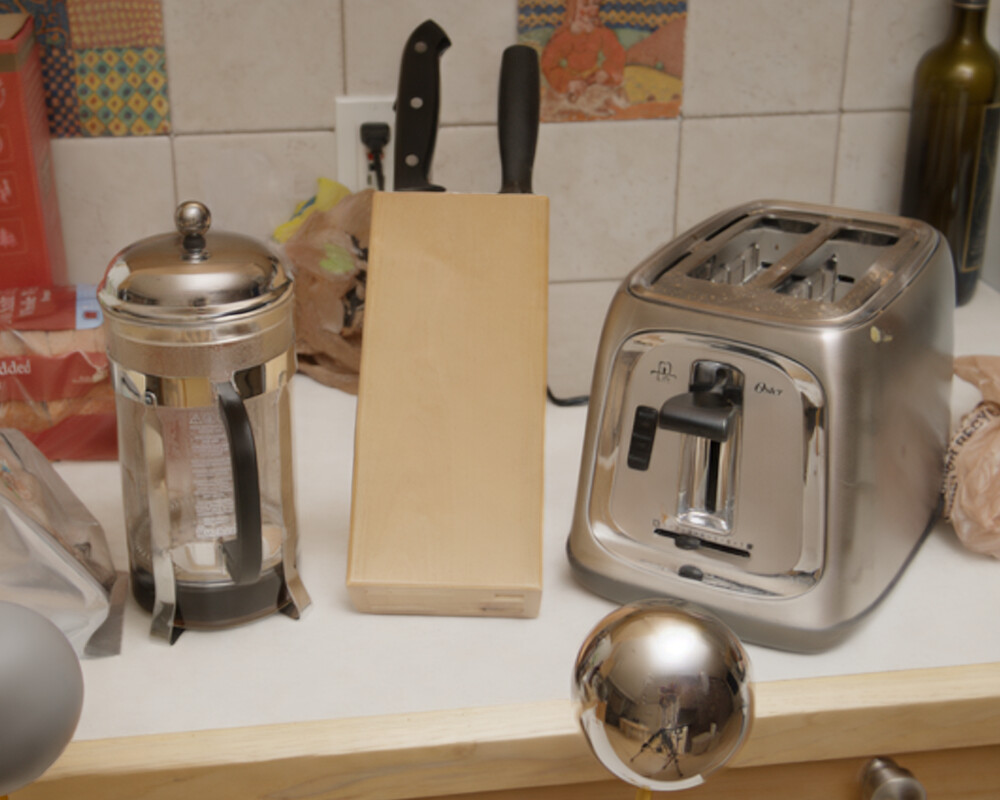} & 
    \includegraphics[width=0.248\linewidth]{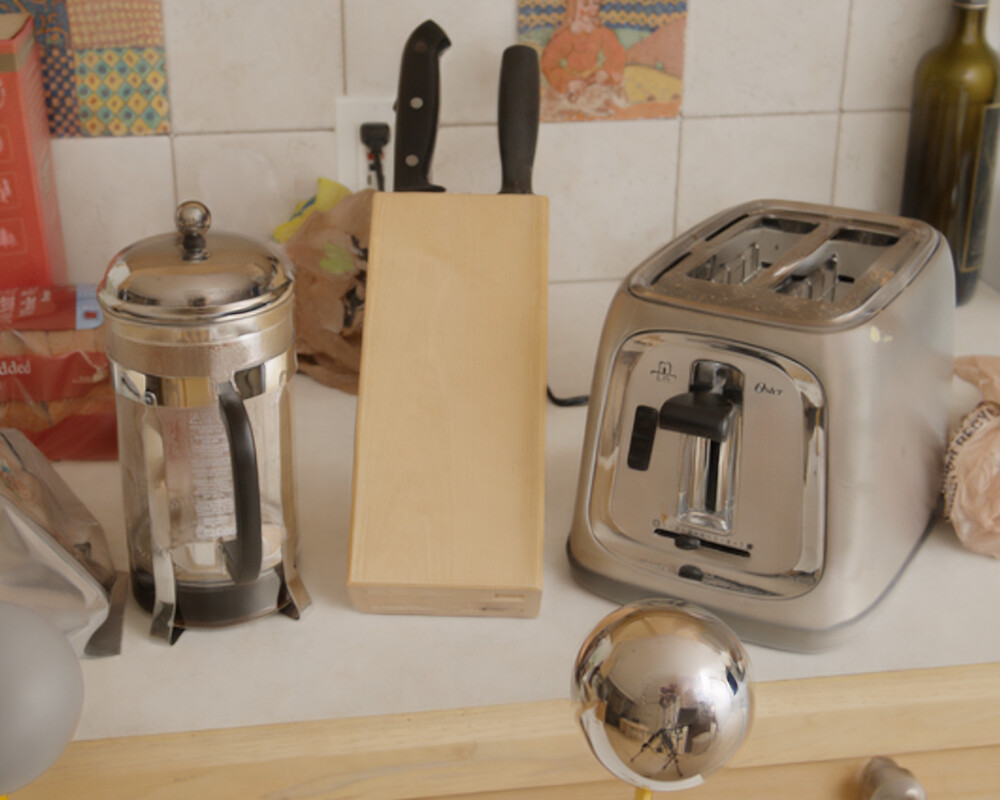} \\
    \includegraphics[width=0.248\linewidth]{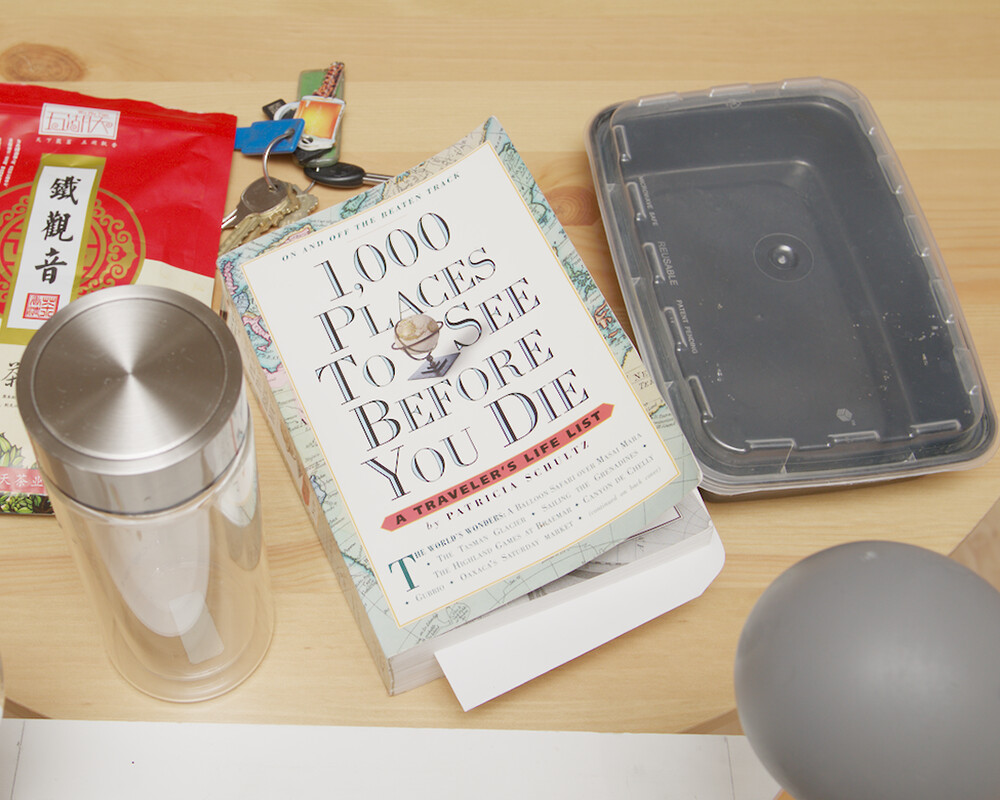} & 
    \setlength{\tabcolsep}{0.25pt}
    \includegraphics[width=0.248\linewidth]{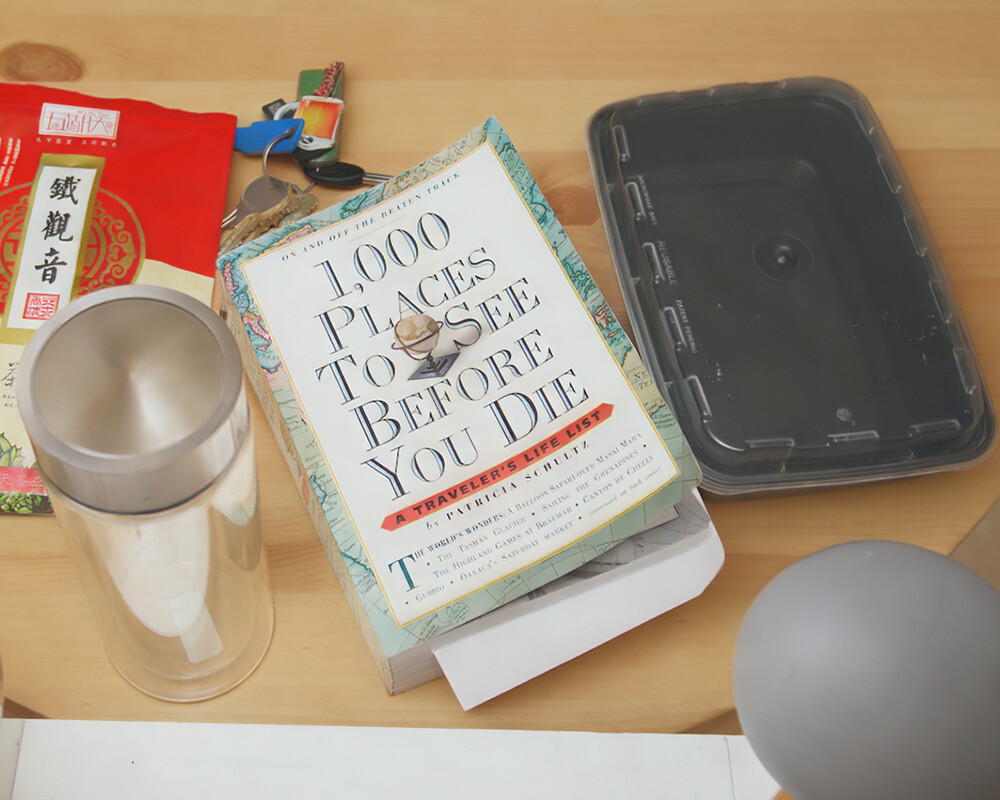} & 
    \includegraphics[width=0.248\linewidth]{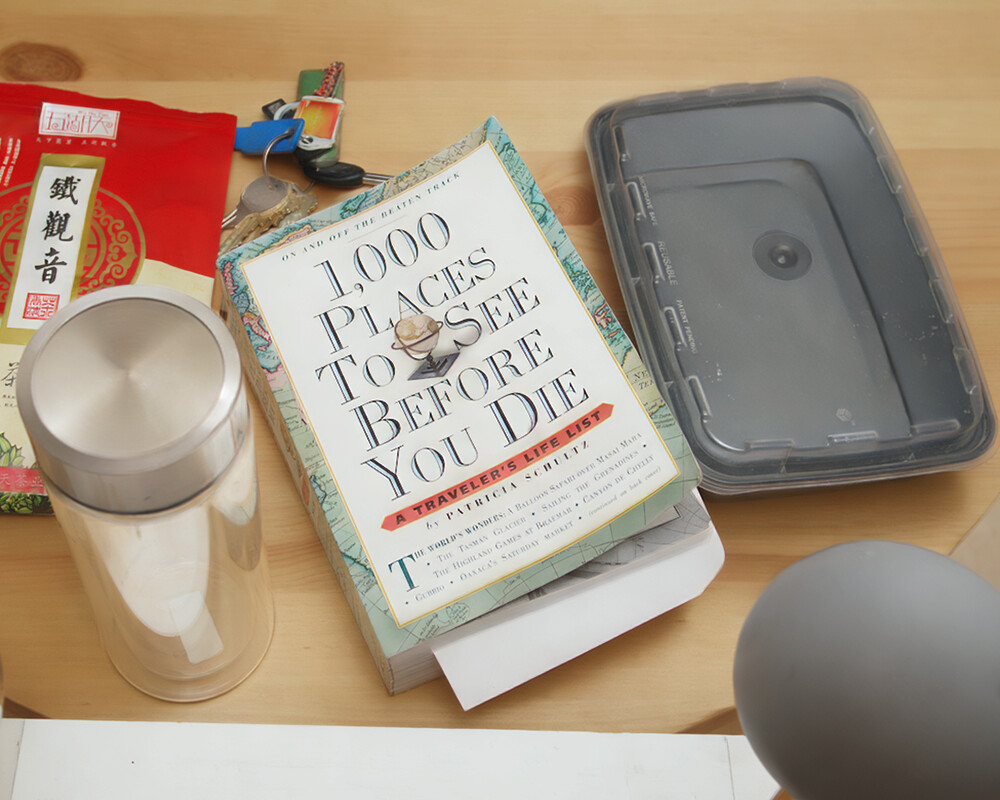} & 
    \includegraphics[width=0.248\linewidth]{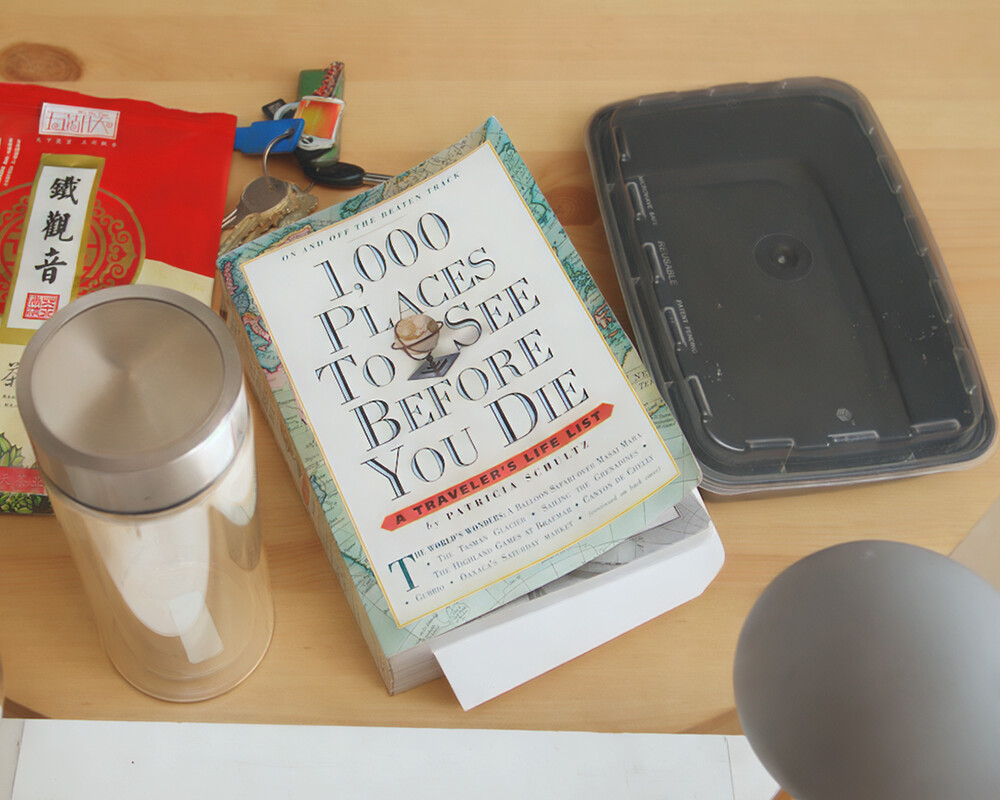} \\ 

    \end{tabular}
    \caption{Relighting results with our light-conditioned ControlNet. From a single input image (left column), the ControlNet can generate realistic relit versions for different target light directions (other columns). Please notice realistic changes in highlights for different light directions \NEW{(top row)}, as well as the synthesis of cast shadows \NEW{(bottom row)}.}
    \label{fig:results-controlnet-2d}
\end{figure*}

Example relighting results obtained using our 2D relighting network on images outside of the dataset are shown in Fig.~\ref{fig:results-controlnet-2d}. Observe how the relit images produced by our method are highly realistic and light directions are consistently reproduced across scenes. A naive solution for radiance field relighting would be to apply this 2D network to each synthesized novel view. However, the ControlNet is not multi-view consistent, and such a naive solution results in significant flickering. Please see the accompanying video for a clear illustration.

\begin{figure*}[!h]
    \includegraphics[width=\linewidth]{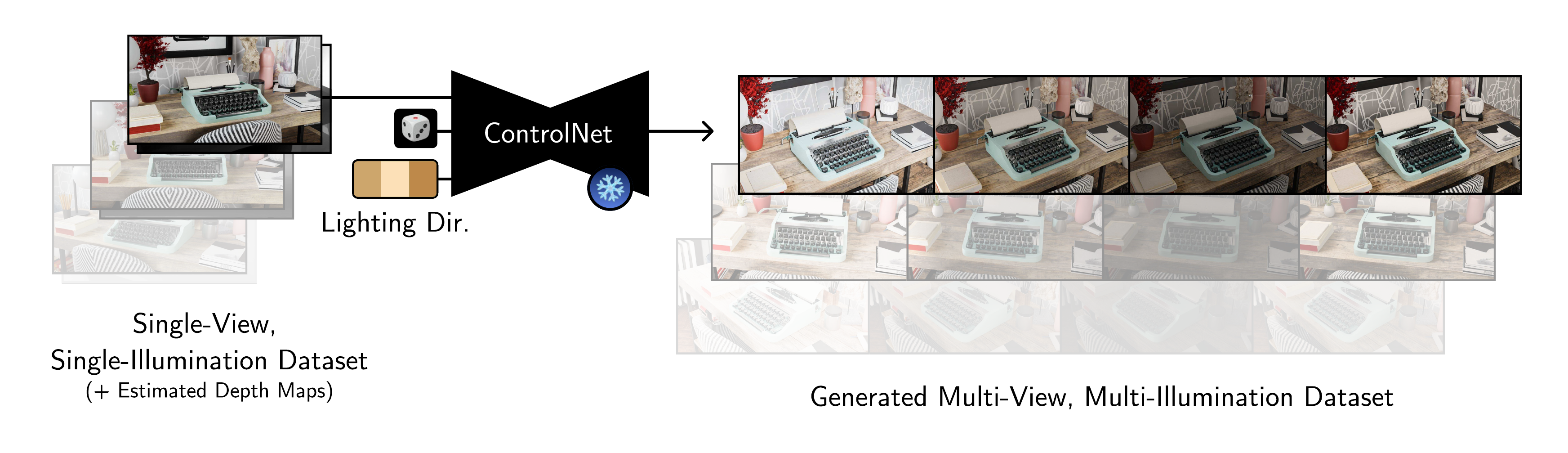}
    \caption{Given a multi-view, single-illumination dataset we use our relighting ControlNet to generate a multi-view, multi-illumination dataset.}
    \label{fig:dataset_xform}
\end{figure*}

\subsection{Augmenting Multi-View/Single-Lighting Datasets}
\label{sec:augmenting}


Given a multi-view set $\mathcal{I}$ of images of a scene captured under the same lighting (suitable for training a radiance field model), we now leverage our light-conditioned ControlNet model to synthetically relight each image in $\mathcal{I}$. We assume the 3D pose of each image $\mathbf{I}_i \in \mathcal{I}$ is known a priori, for example via Colmap~\cite{colmap1,colmap2}. We then simply relight each $\mathbf{I}_i \in \mathcal{I}$ to the corresponding 18 known light directions in the dataset from Murmann et al.~\shortcite{multilum} (excluding the directions where the flash points forward), (see Sec.~\ref{sec:2d-relight}). We now have a full multi-lighting, multi-view dataset. This process is illustrated in Fig.~\ref{fig:dataset_xform}.

\subsection{Training a Lighting-Consistent Radiance Field}
\label{sec:correcting}

Given the generated multi-light, multi-view dataset, we now describe our solution to provide a relightable radiance field. In particular, we build on the 3DGS framework of Kerbl et al.~\shortcite{gaussian-splats}. 
Our requirements are twofold: first, define an augmented radiance field that can represent lighting conditions from different lighting directions; second, allow direct control of the lighting direction used for relighting.


The original 3DGS~\cite{gaussian-splats} radiance field uses spherical harmonics (SH) to represent view-dependent illumination. To encode varying illumination, we replace the SH coefficients with a 3-layer MLP $\boldsymbol{c}_\theta$ of width 128 \NEW{which takes as input the light direction along with the viewing direction. Both vectors have a size of 16 after encoding.} 

Since light directions are computed with respect to a local camera reference frame (c.f. Sec.~\ref{sec:2d-relight}), we subsequently register them to the world coordinate system (obtained from Colmap) by rotating them according to their (known) camera rotation parameters:
\begin{equation}
\boldsymbol{l}' = \boldsymbol{R}_i\boldsymbol{l} \,,
\label{eq:light-rotation}
\end{equation}
where $\boldsymbol{R}_i$ is the $3\times3$ camera-to-world rotation matrix of image $\mathbf{I}_i$ from its known pose.

We condition the MLP with the spherical harmonics encoding of the globally consistent lighting direction $\boldsymbol{l}'$, which enables training a 3DGS representation on our multi-lighting dataset. While this strategy works well for static images, it results in inconsistent lighting across views despite accounting for camera rotation in Eq.~\ref{eq:light-rotation}. \NEW{Radiance fields like 3DGS rely on multi-view consistency, and breaking it introduces additional floaters and holes in surfaces.}


To allow the neural network to account for this inconsistency and correct accordingly, we optimize a per-image auxiliary latent vector $\boldsymbol{a}$ \NEW{of size 128}. Similar approaches for variable appearance have been used for NeRFs~\cite{nerfw}. 
Therefore, in addition to the lighting direction $\boldsymbol{l}'$, we condition the MLP with per-view auxiliary parameters $\boldsymbol{a}$: 
%
\begin{equation}
\boldsymbol{c}(\boldsymbol{o}, \boldsymbol{d}) = \sum_{g=1}^G w_g\boldsymbol{c}_{\theta}(\boldsymbol{x}_g, \boldsymbol{d}|\boldsymbol{a}_{v},\boldsymbol{l}') \,,
\end{equation}
where $g \in [1, G]$ sums over the $G$ gaussians (see \cite{gaussian-splats}), $\boldsymbol{x}_g/w_g$ are their features/weights, $\boldsymbol{d}$ is the view direction, $\boldsymbol{o}$ the ray origin, and $\boldsymbol{c}$ is the predicted pixel color. Note that for novel views at inference, we use as latent vector the mean of all training view latents i.e. $\mathbb{E}_v[\boldsymbol{a}_v]$.

%

We first train 3DGS with the unlit images as a ``warmup'' stage for 5K iterations, then train the full multi-illumination solution for another 25K iterations, using all 18 back-facing light directions (see Sec.~\ref{sec:2d-relight}). The multi-illumination nature of the training results in an increase in ``floaters''. As observed by Philip and Deschaintre~\cite{floatersnomore}, floaters are often present close to the input cameras; the explicit nature of 3DGS allows us to reduce these effectively. In particular, we calculate a $z_{near}$ value for all cameras by taking the $z$ value of the 1st percentile of nearest SfM points and scaling this value down by 0.9. During training, at each step, all gaussian primitives that project within the view frustum of a camera but are located in front of its $z_{near}$ plane are culled. Finally, given the complexity of modeling variable lighting, we observed that the optimization sometimes converges to blurry results. To counter this, we overweight three front-facing views (left, right, and center), by optimizing for one of these views every three iterations. This provides marginal improvement in results; all images shown are computed with this method, but it is optional.

\begin{figure}[t]
    \includegraphics[width=\linewidth]{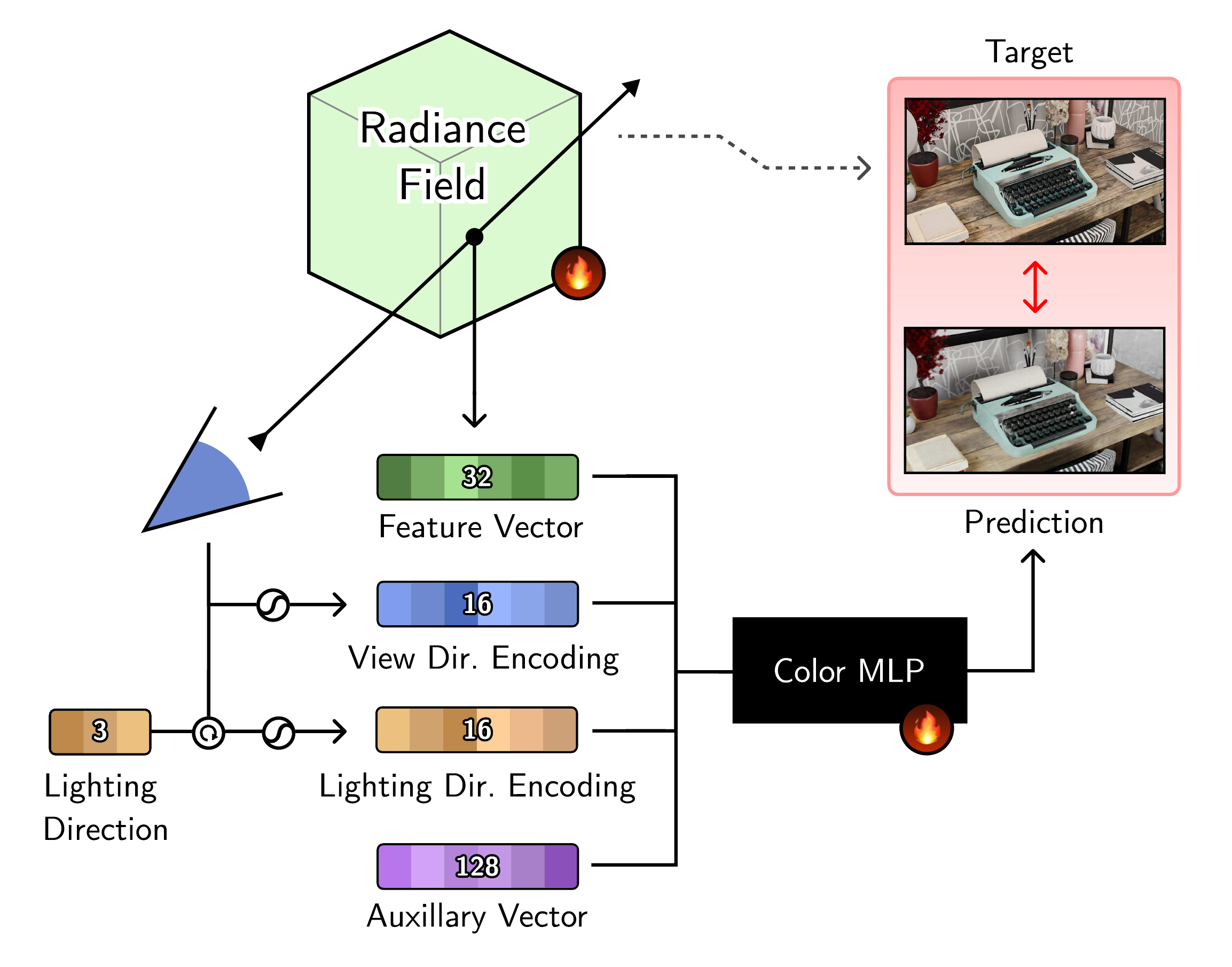}
    \caption{\NEW{Overview of our radiance field training scheme. To alleviate potential inconsistencies in lighting directions, we condition our 3DGS-based radiance field both on the illumination direction encoding and on optimized auxiliary vectors (one per training image). These vectors model the differences between predictions and let us fit each view to convergence.}}
    \label{fig:method-3d}
\end{figure}

The full method for relightable radiance fields is shown in Fig.~\ref{fig:method-3d}. At inference, we can directly choose a lighting direction, and use efficient 3DGS rendering for \REALTIME updates with modified lighting. \NEW{Our latent vectors and floater removal remove most, but not all, artifacts introduced by the multi-view inconsistencies; this can be seen in the ablations at the end of the supplemental video.}


\begin{figure*}[!h]
    \setlength{\tabcolsep}{1pt}

    \begin{tabular}{ccc@{\hskip 0.3cm}ccc}
        \raisebox{53pt}{\includegraphics[width=0.04\textwidth,raise=-\dp\strutbox]{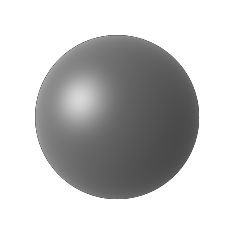}} &
        \includegraphics[width=0.215\linewidth,raise=-\dp\strutbox]{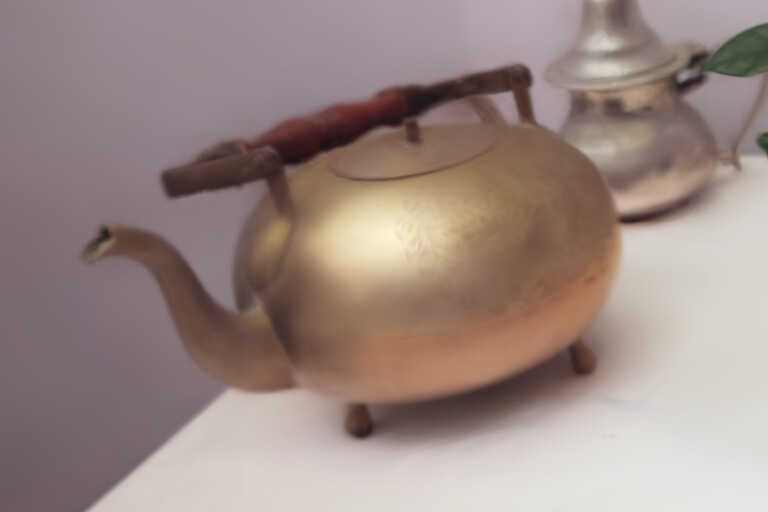} &
        \includegraphics[width=0.215\linewidth,raise=-\dp\strutbox]{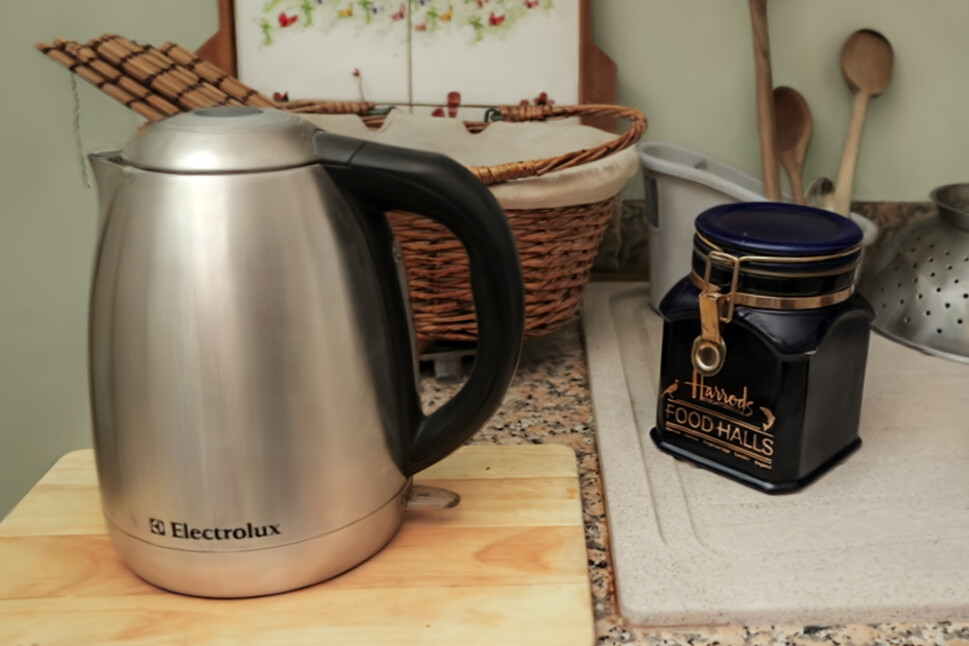} &
        
        \raisebox{53pt}{\includegraphics[width=0.04\textwidth,raise=-\dp\strutbox]{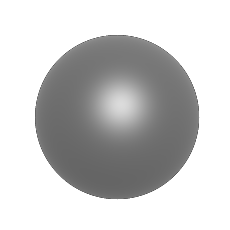}} &
        \includegraphics[width=0.215\linewidth,raise=-\dp\strutbox]{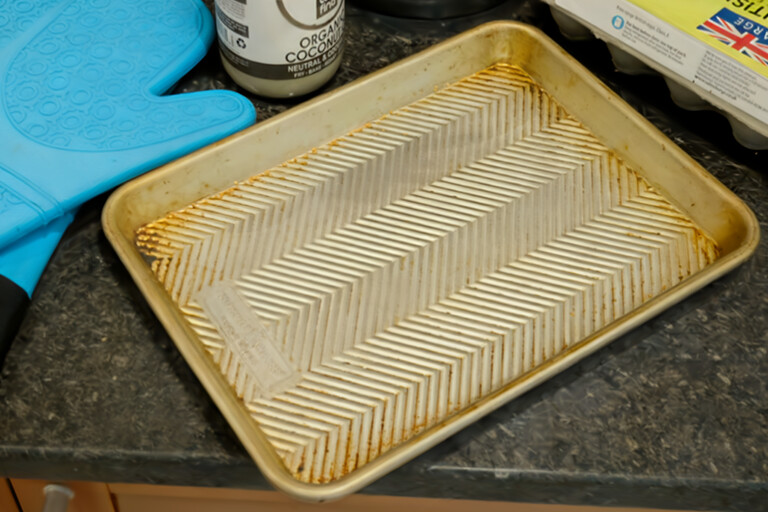} &
        \includegraphics[width=0.215\linewidth,raise=-\dp\strutbox]{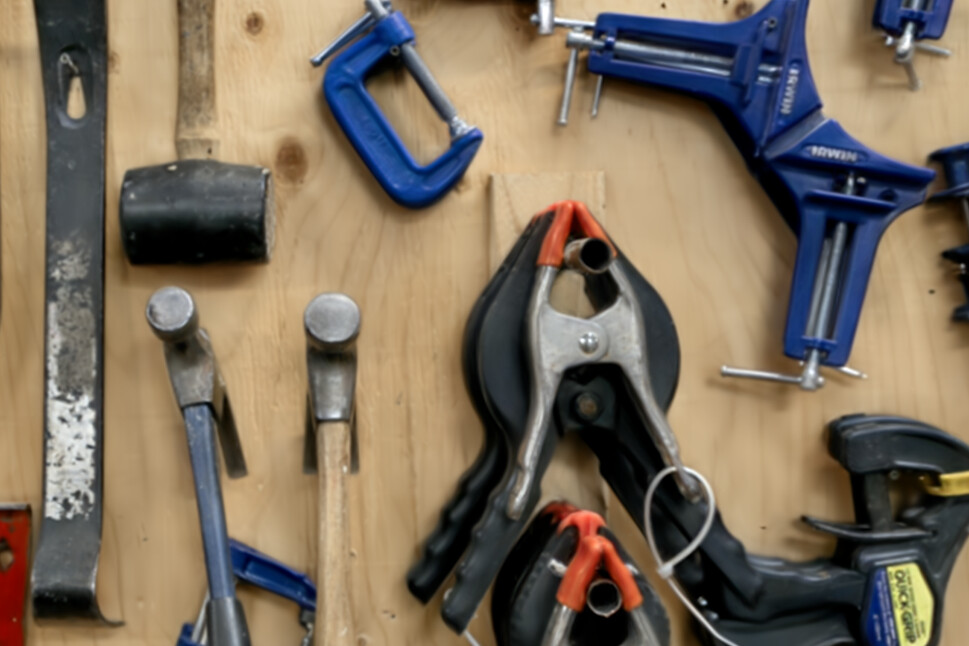} \\

        \raisebox{53pt}{\includegraphics[width=0.04\textwidth,raise=-\dp\strutbox]{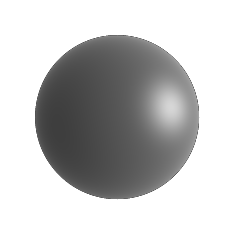}} &
        \includegraphics[width=0.215\linewidth,raise=-\dp\strutbox]{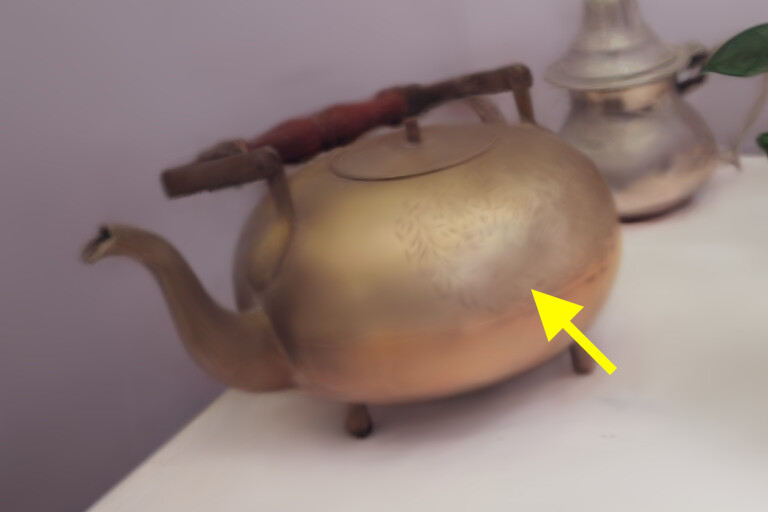} &
        \includegraphics[width=0.215\linewidth,raise=-\dp\strutbox]{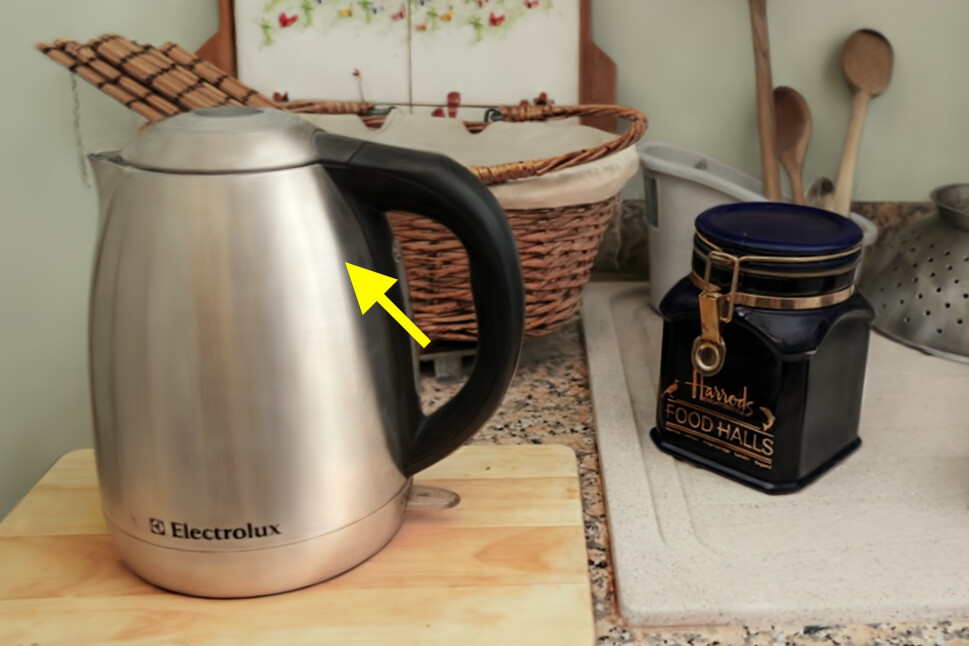} &

        \raisebox{53pt}{\includegraphics[width=0.04\textwidth,raise=-\dp\strutbox]{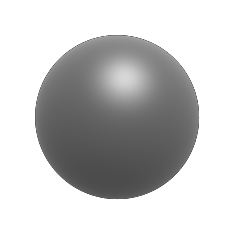}} &
        \includegraphics[width=0.215\linewidth,raise=-\dp\strutbox]{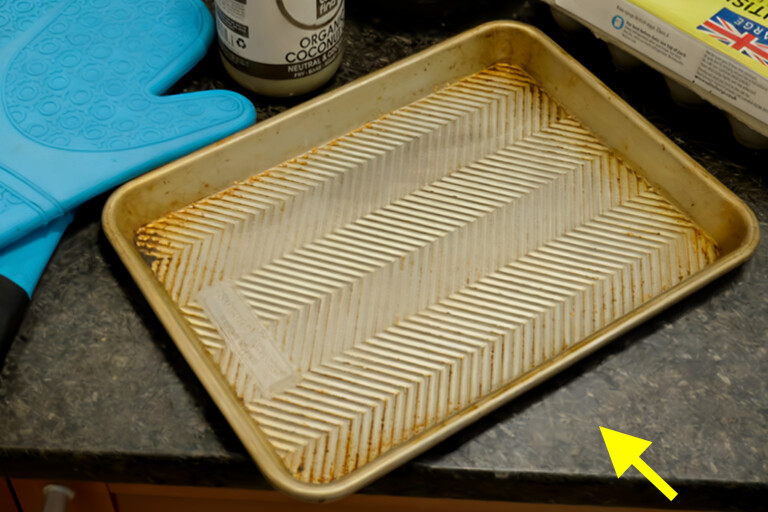} &
        \includegraphics[width=0.215\linewidth,raise=-\dp\strutbox]{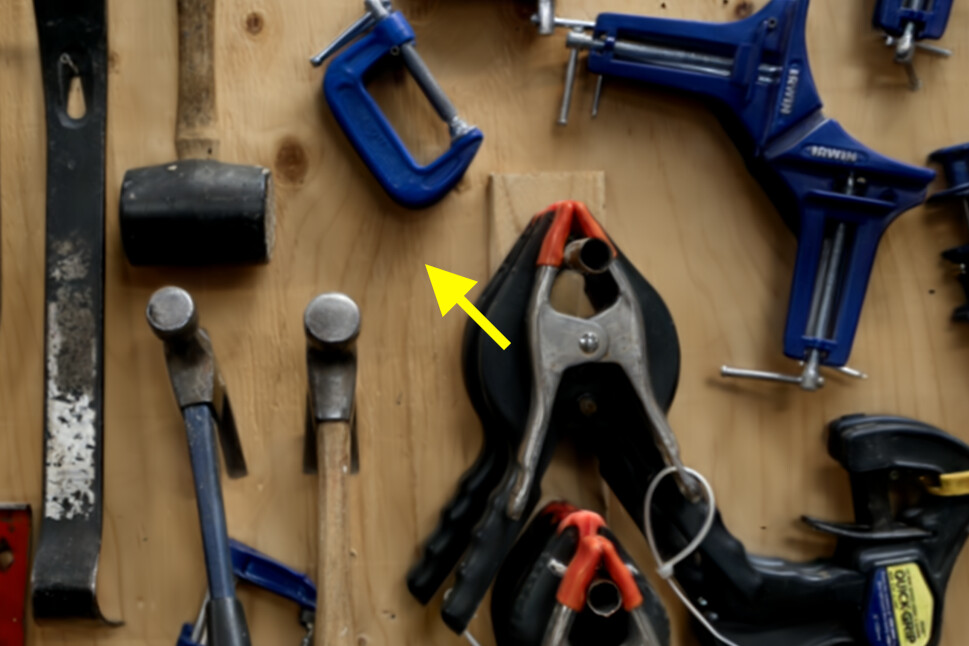} \\
        


    \end{tabular}
    
    \vspace{0.5em}
    \caption{
    \label{fig:results-3d-real}
    \NEW{Qualitative relighting results for the real scenes, from left to right: \textsc{Chest of Drawers}, \textsc{Kettle},  \textsc{MipNeRF Room} and \textsc{Garage Wall}, for a moving light source. The lighting direction is indicated in the gray ball in the lower right. Please see the supplemental video for more results. Please note how the highlights (left group) and shadows (right group) have changed.}
    }
\end{figure*}
\section{Results and Evaluation}

Our method was implemented by leveraging publicly available implementations of ControlNet~\cite{controlnet} and 3DGS~\cite{gaussian-splats}. We use Stable Diffusion~\cite{stable-diffusion} v2.1 as a backbone. Our source code and datasets will be released upon publication.

We first present the results of our 3D relightable radiance field, both for synthetic and real-world scenes. We then present a quantitative and qualitative evaluation of our method by comparing it to previous work and finally present an ablation of the auxiliary vector $a$ from Sec.~\ref{sec:correcting}.

\subsection{Test Datasets}

Since there are no real \emph{multi-view} multi-illumination indoor datasets of full scenes available for our evaluation, we use synthetic scenes to allow quantitative evaluation. For this purpose, we designed 4 synthetic test scenes (\textsc{kitchen}, \textsc{livingroom}, \textsc{office}, \textsc{bedroom}). They were created in Blender by downloading artist-made 3D rooms from Evermotion and modifying them to increase clutter: in each room, we gathered objects and placed them on a table or a countertop. 
We also created simpler, diffuse-only versions to evaluate how scene clutter affects the relighting results.
For each synthetic scene, we first built a standard multi-view (single-lighting) dataset consisting of 4 camera sweeps (left-to-right, at varying elevations) of 50 frames for training and one (at a different elevation) of 100 frames for testing. We simulated the light direction of the 2D training dataset with a spotlight with intensity of 2 kW and radius 0.1 locating on top of the camera and pointing away from it. We used the same set of camera flash directions as in the dataset of Murmann et al.~\shortcite{multilum}. We then render all frames in $736 \times 512$ using the Cycles path tracer. Please note that the effective lighting direction will be dependent on the exact configuration of the room. This configuration is our best effort to produce a ground truth usable for comparison.

In addition, we also captured a set of real scenes (\textsc{Kettle}, \textsc{Hot Plates}, \textsc{Paint Gun}, \textsc{Chest of Drawers} and \textsc{Garage Wall}), for which we performed a standard radiance-field multi-view capture, by taking between $90$--$150$ images of the environment, in an approximate sphere (or hemisphere) around the scene center of interest.


\subsection{3D Relighting Results}

We begin by showing qualitative results on the set of real scenes that we captured. Here, we used a resolution of $1536 \times 1024$, training for 150K iterations. We show qualitative results for these scenes using our 3D relightable radiance field in Fig.~\ref{fig:results-3d-real}. In addition, we also show results for two scenes from the MipNeRF360 dataset, namely \textsc{Counter} and \textsc{Room}. 

As our method is lightweight and only adds a small MLP over the core 3DGS architecture, it runs \INREALTIME for both novel view synthesis and relighting at 30fps on an A6000 GPU. \NEW{Memory usage is comparable to the original 3DGS.} Please see the video for \REALTIME relighting results on these scenes and additional synthetic scenes. We see that our method produces realistic and plausible relighting results. Also, note that our solution is temporally consistent.

\begin{figure*}[!h]
    \includegraphics[width=\linewidth]{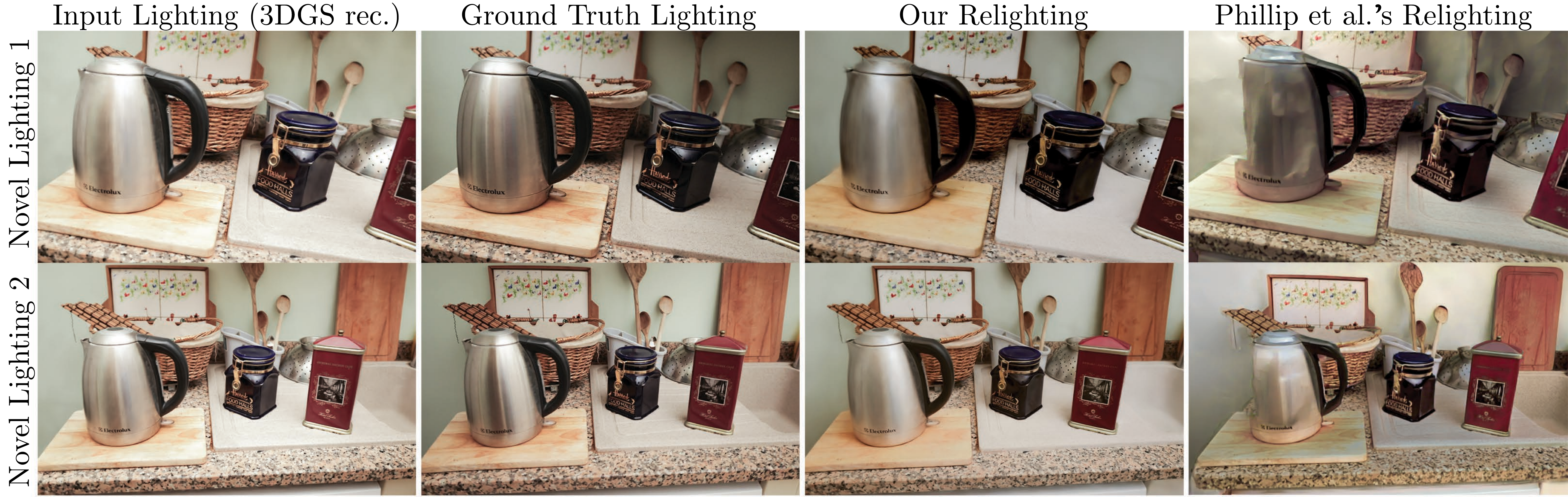}

    \caption{Qualitative comparison on real scene \textsc{Kettle}. From left to right, from the same viewpoint: input lighting condition (view reconstructed using 3D Gaussian Splatting), target lighting, our relighting, Philip et al.~\cite{philip2021free} relighting. Top and bottom rows are two different lighting conditions. Philip et al.~\cite{philip2021free} exhibits much more geometry and shading artifacts compared to our method; in particular imprecise MVS preprocessing results in missing geometry.}
    \label{fig:real-comp-qualitative}
\end{figure*}

\subsection{Evaluation}




\begin{figure*}[!h]
    \footnotesize
    \setlength{\tabcolsep}{0.24pt}
    \newlength{\mylength}
    \setlength{\mylength}{0.18\textwidth}
    \begin{tabular}{cccccc}
        & GT & Ours & Outcast & R3DG & TensoIR \\  
        \raisebox{123px}{\includegraphics[width=0.04\textwidth]{figures/gray_balls/left.png}} &
        \includegraphics[width=\mylength]{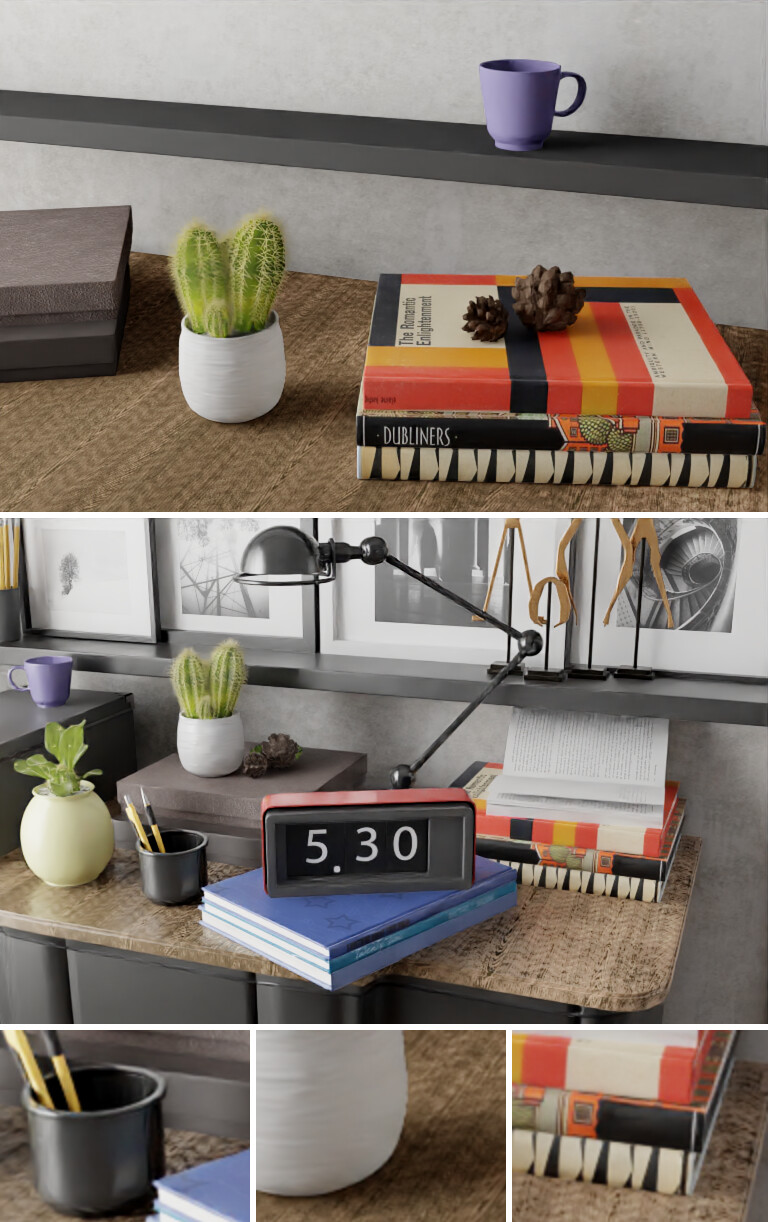} &
        \includegraphics[width=\mylength]{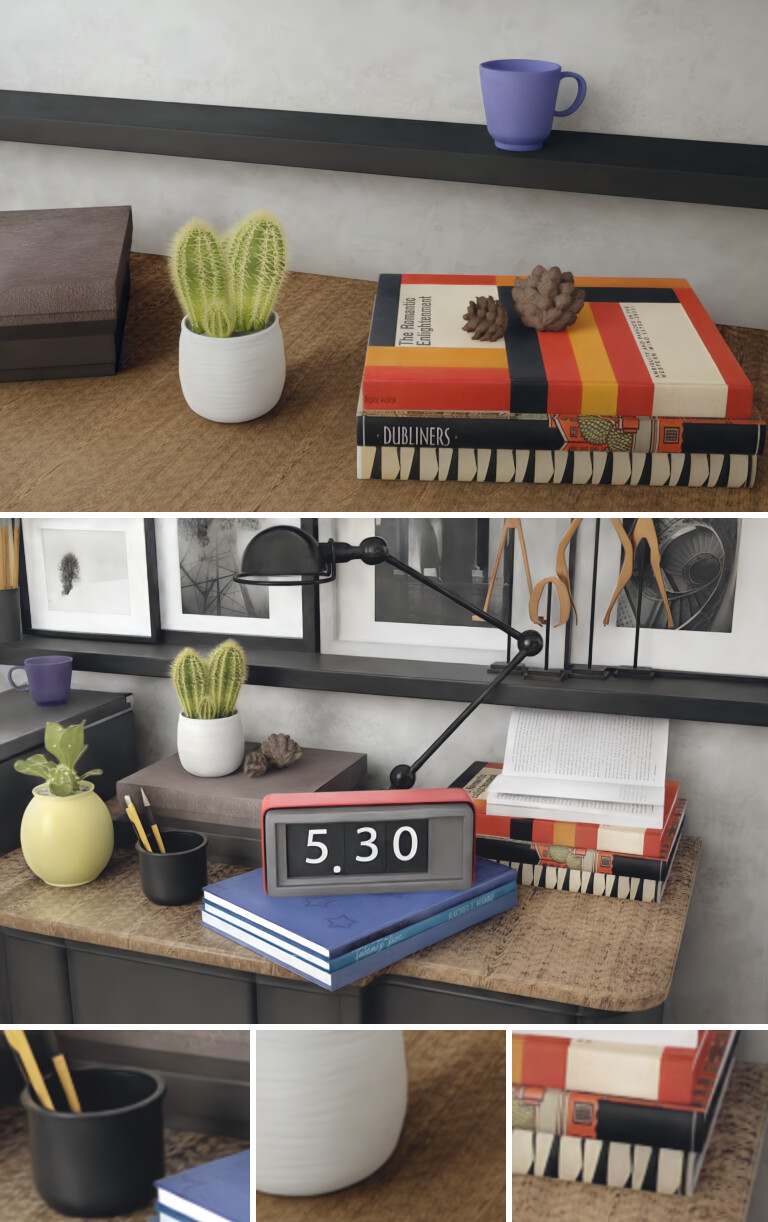} &
        \includegraphics[width=\mylength]{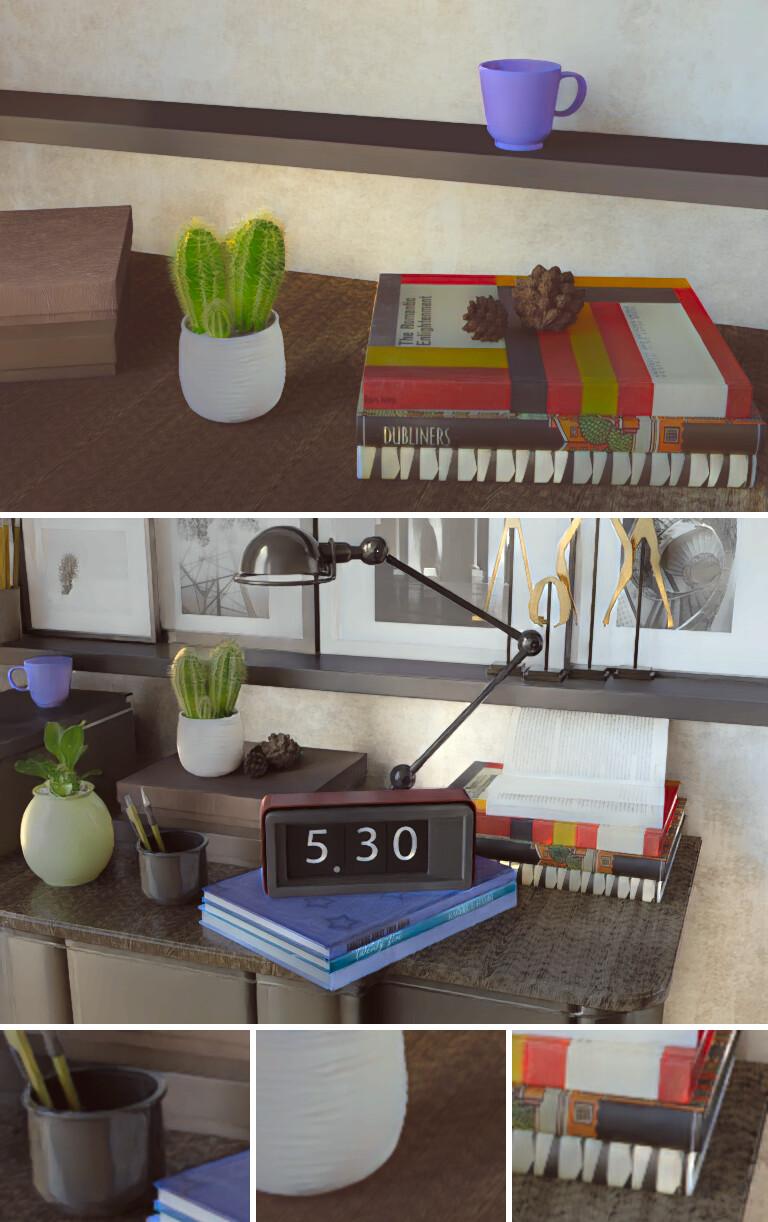} &
        \includegraphics[width=\mylength]{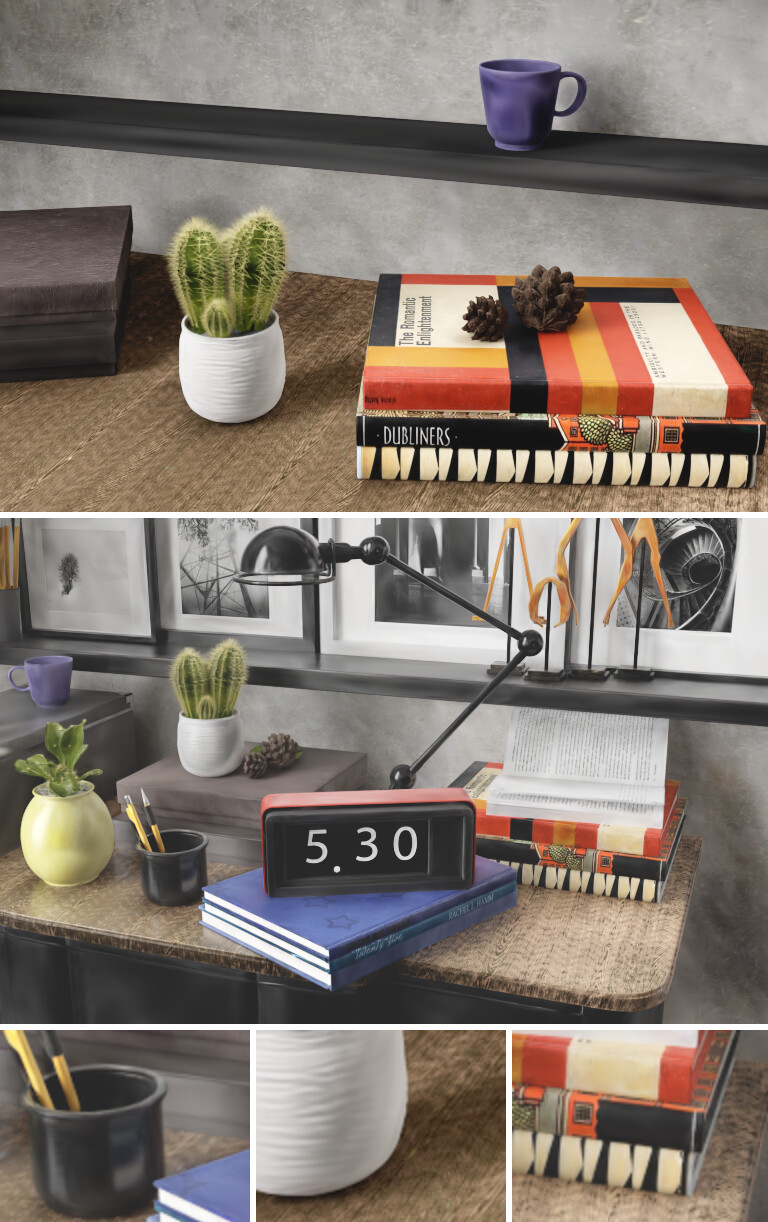} &
        \includegraphics[width=\mylength]{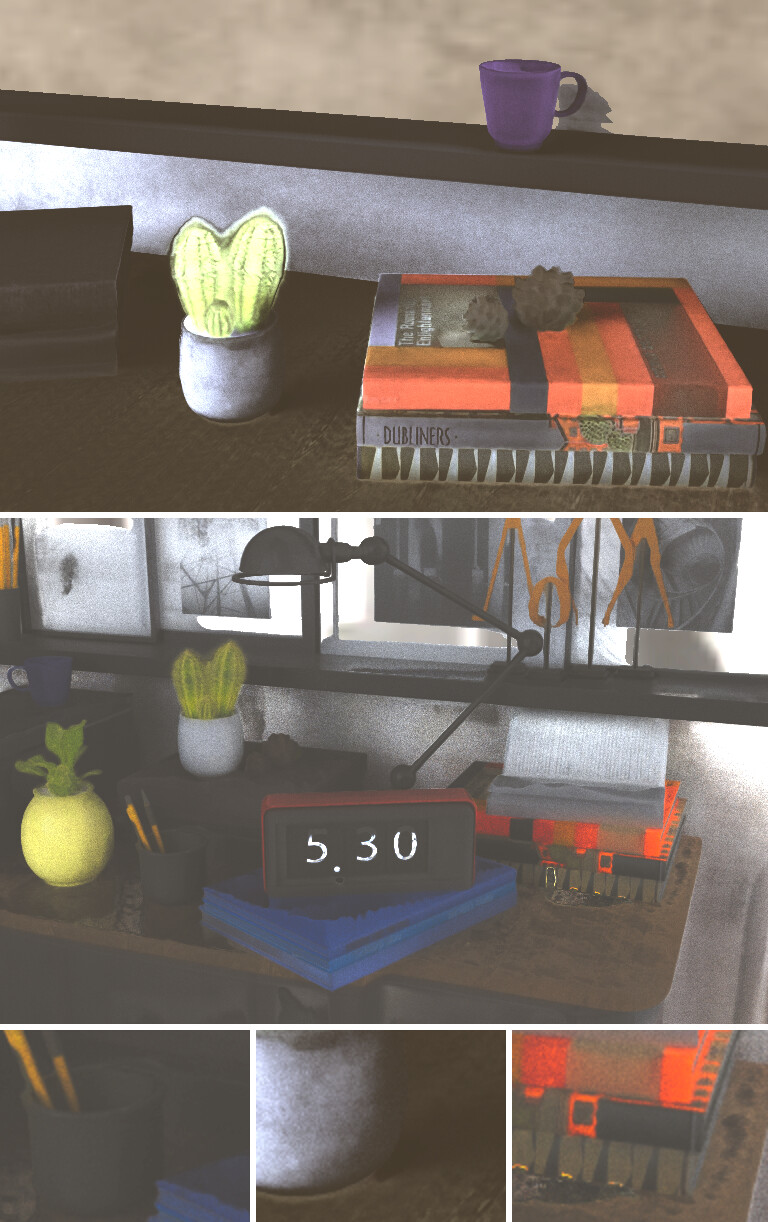} \\
        \raisebox{123px}{\includegraphics[width=0.04\textwidth]{figures/gray_balls/right.png}} &
        \includegraphics[width=\mylength]{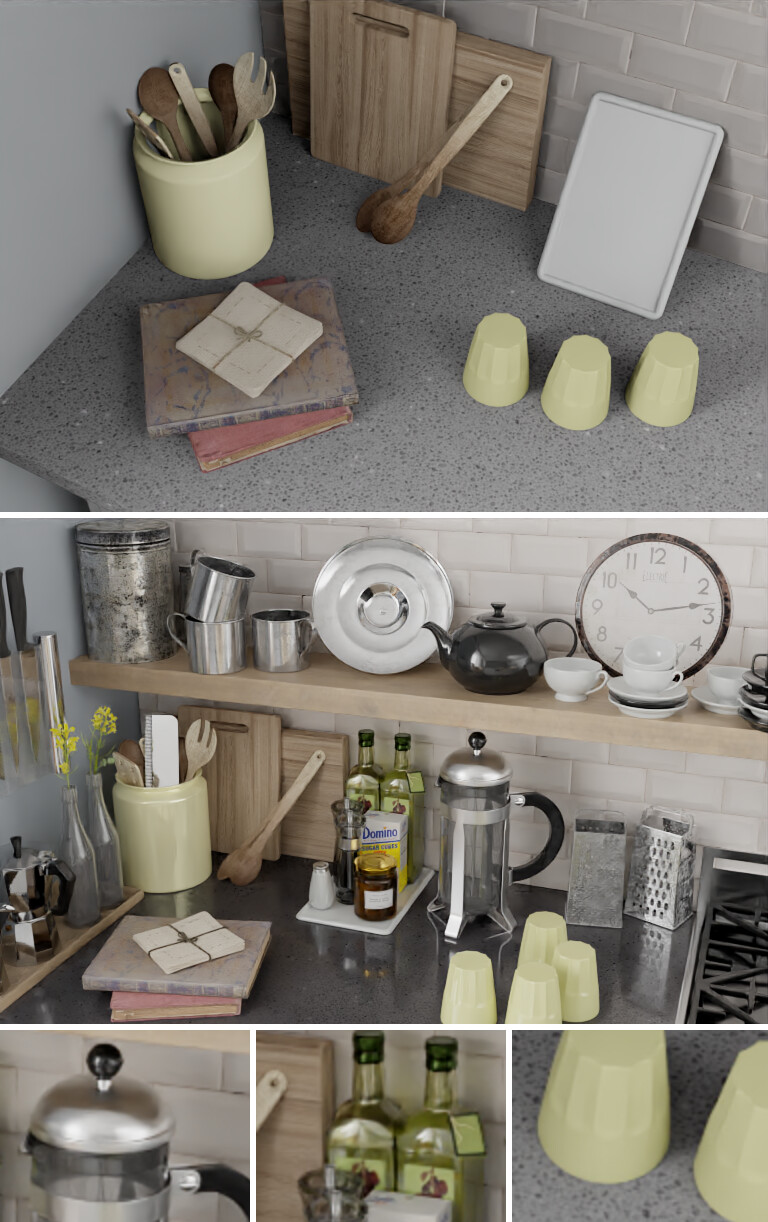} &
        \includegraphics[width=\mylength]{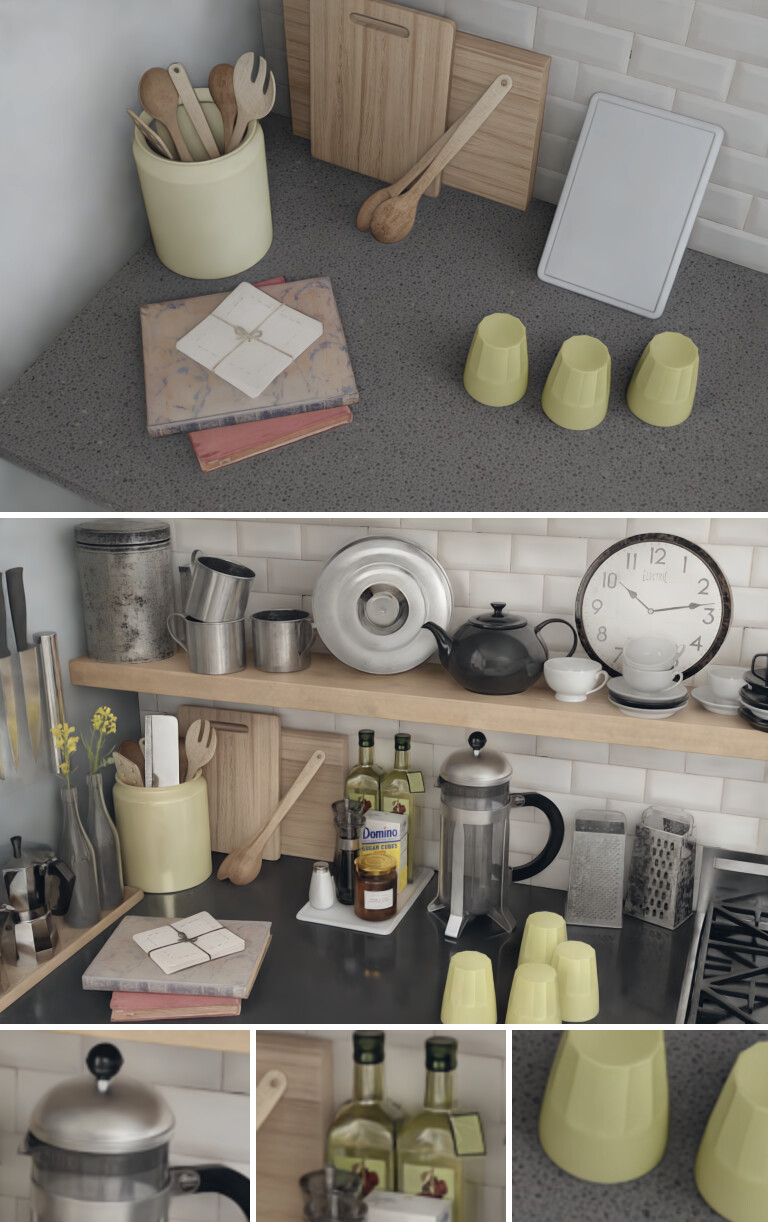} &
        \includegraphics[width=\mylength]{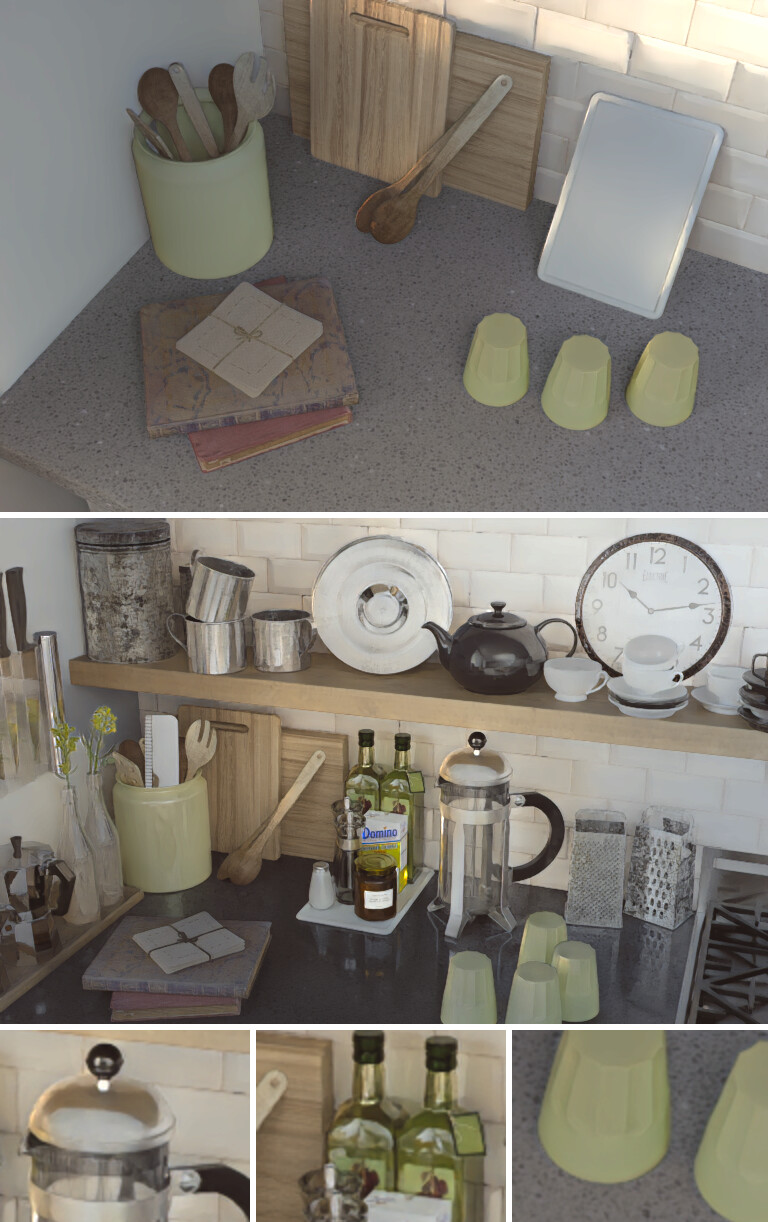} &
        \includegraphics[width=\mylength]{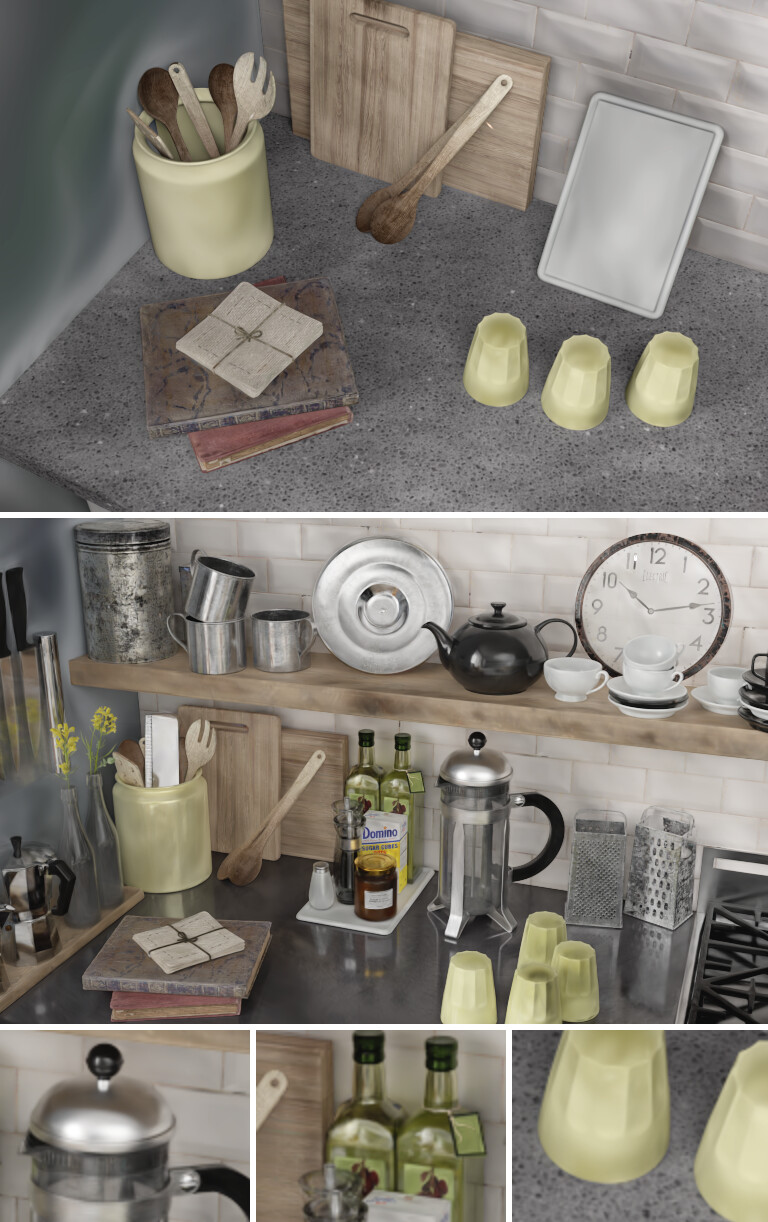} &
        \includegraphics[width=\mylength]{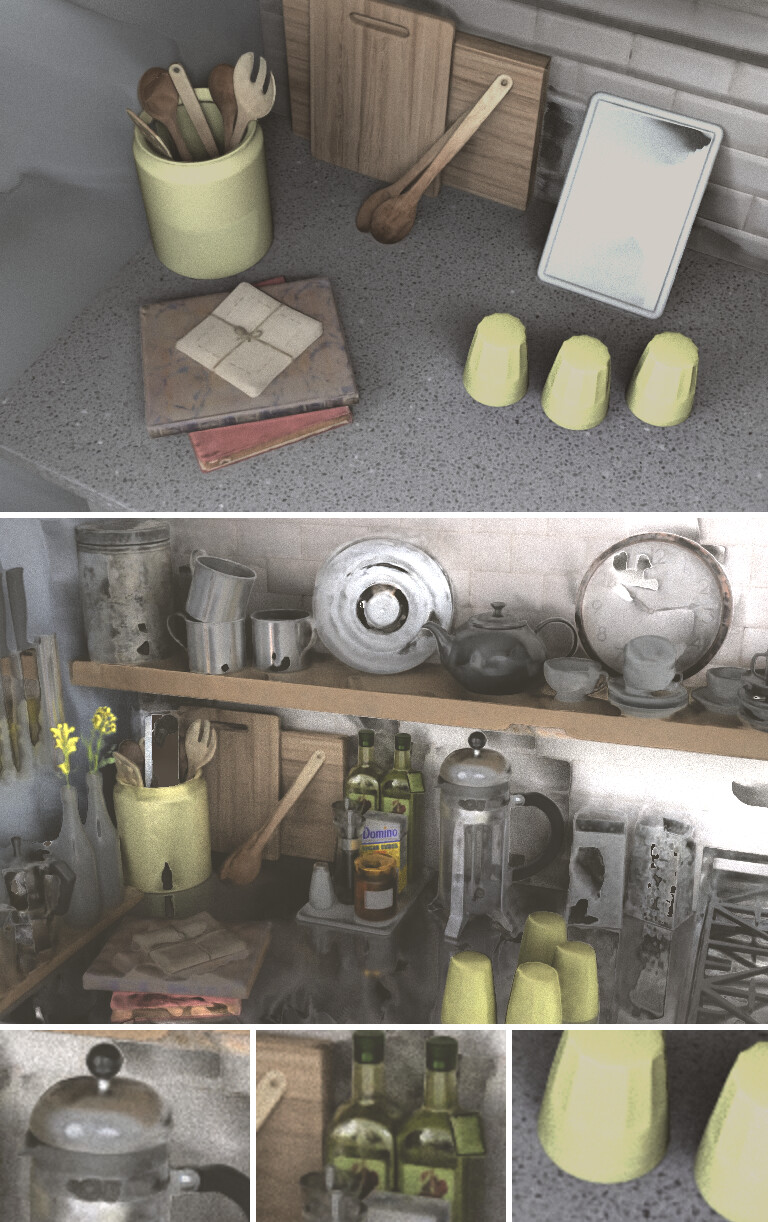} \\
        \raisebox{123px}{\includegraphics[width=0.04\textwidth]{figures/gray_balls/center.png}} &
        \includegraphics[width=\mylength]{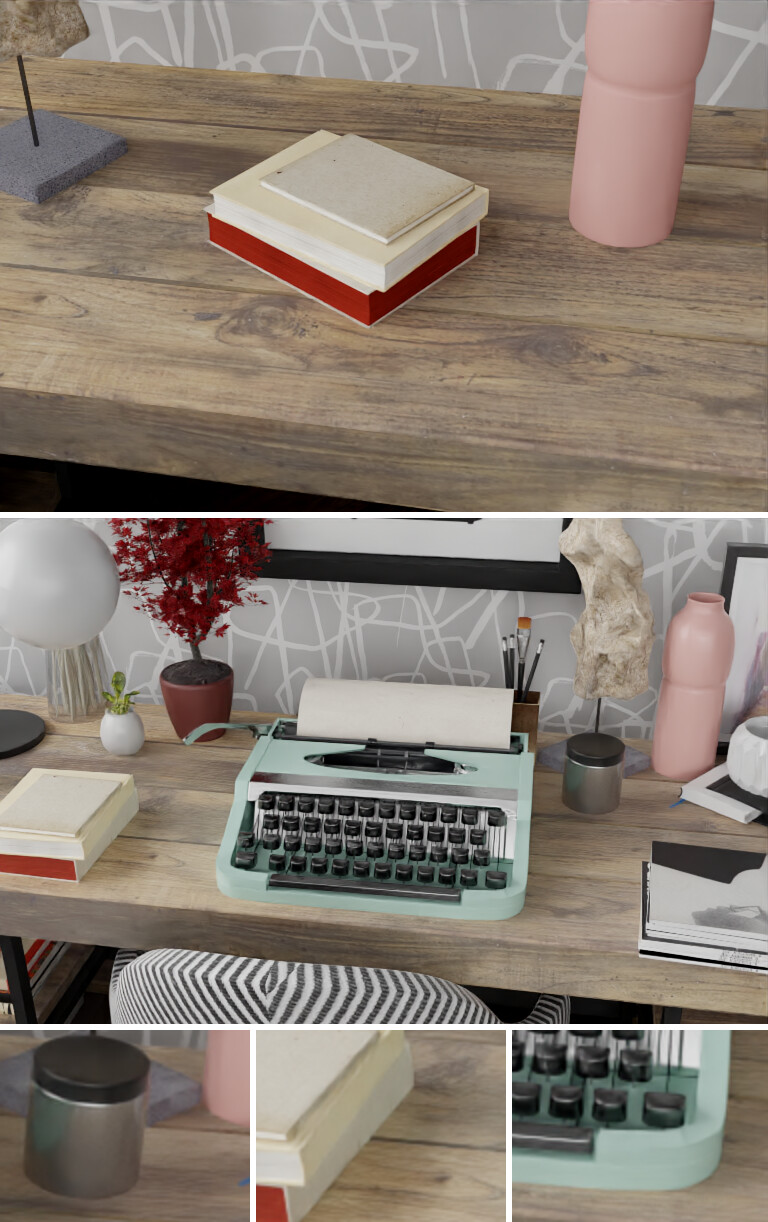} &
        \includegraphics[width=\mylength]{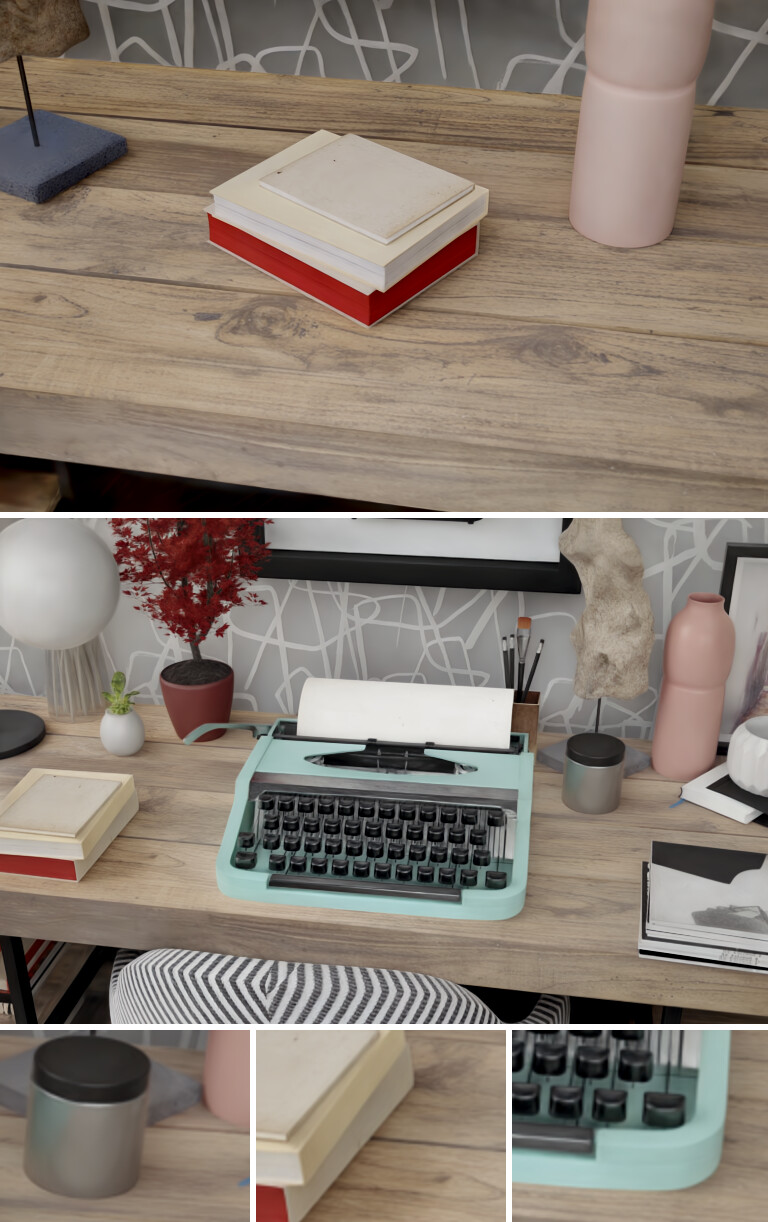} &
        \includegraphics[width=\mylength]{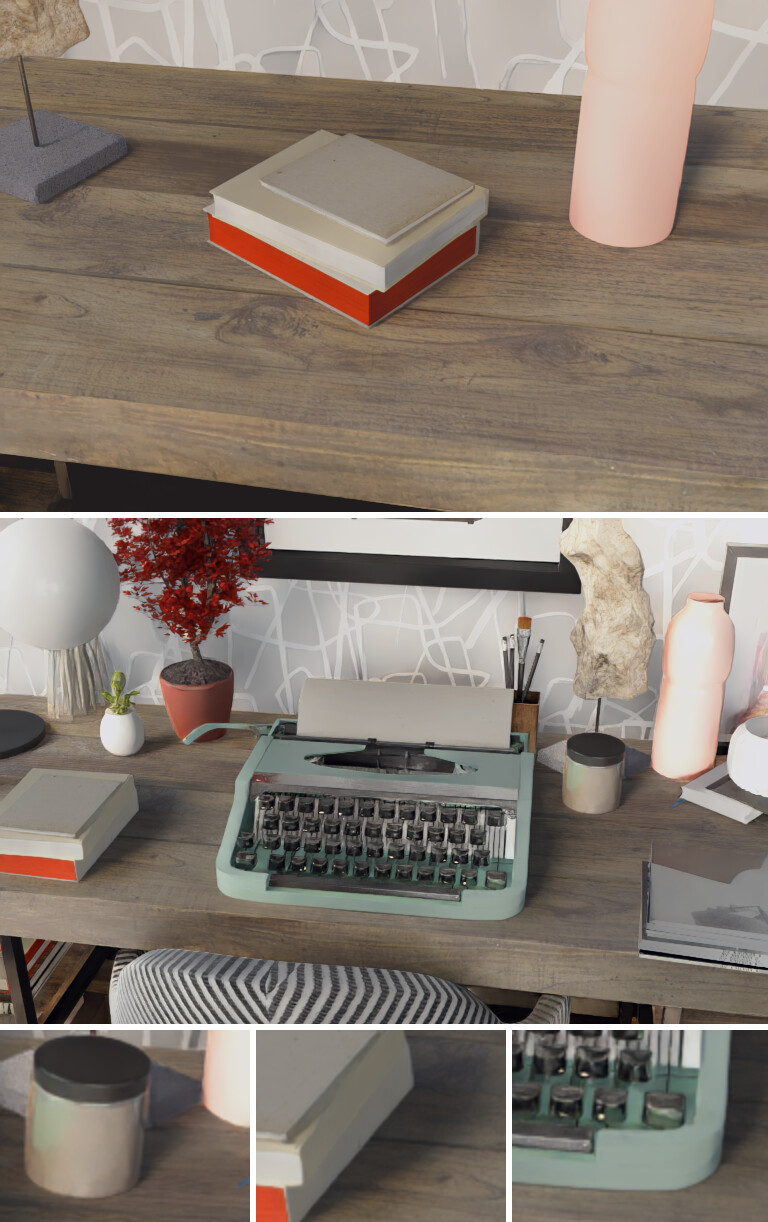} &
        \includegraphics[width=\mylength]{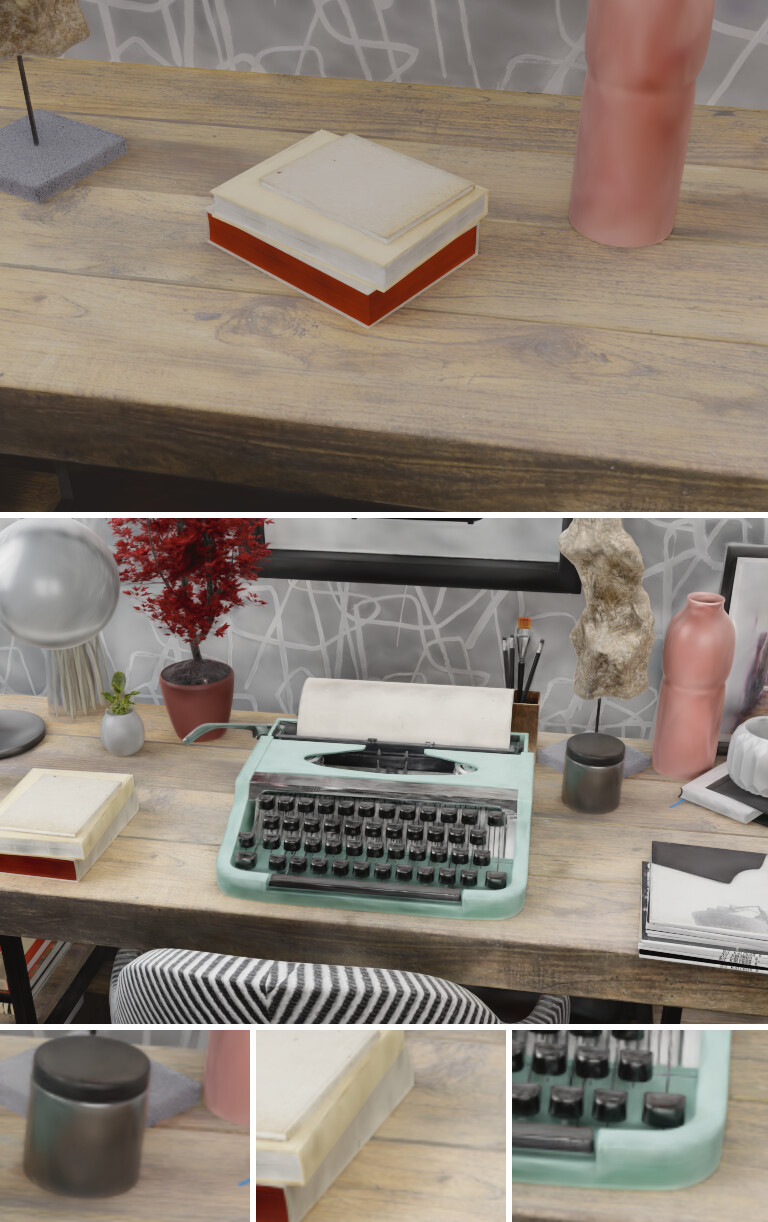} &
        \includegraphics[width=\mylength]{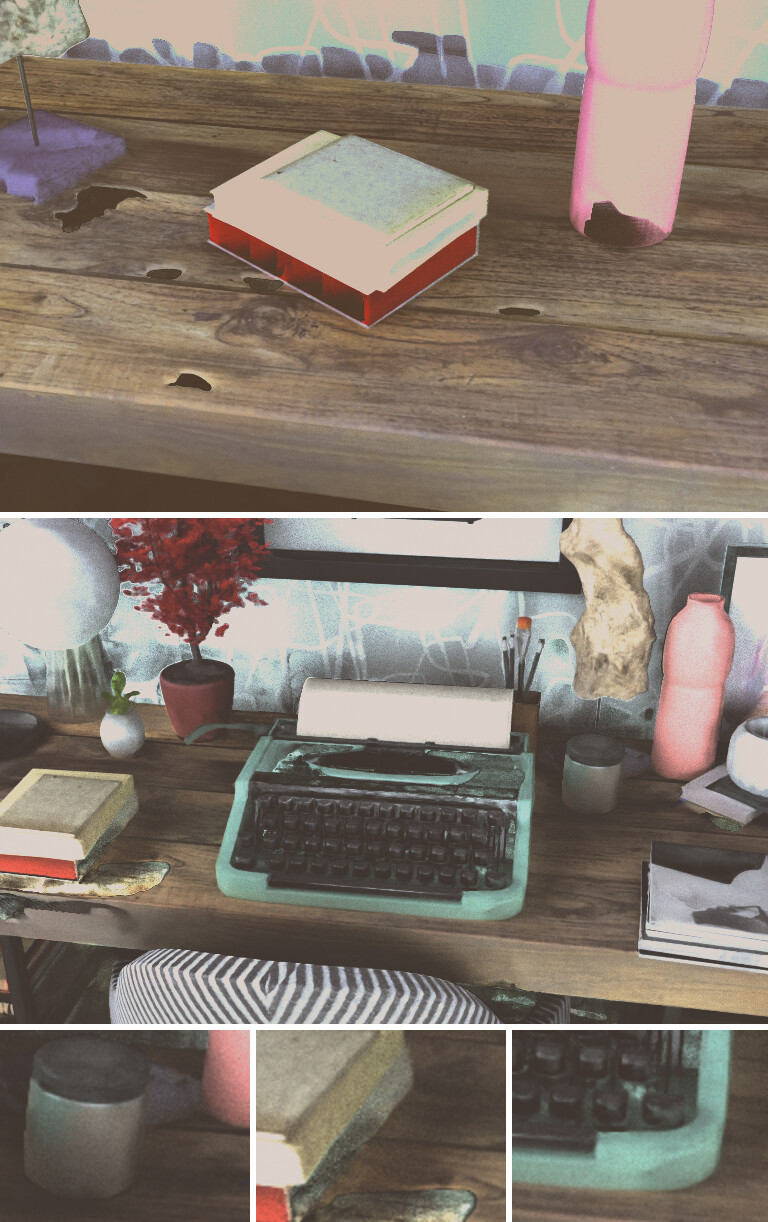} \\
        \raisebox{123px}{\includegraphics[width=0.04\textwidth]{figures/gray_balls/top.png}} &
        \includegraphics[width=\mylength]{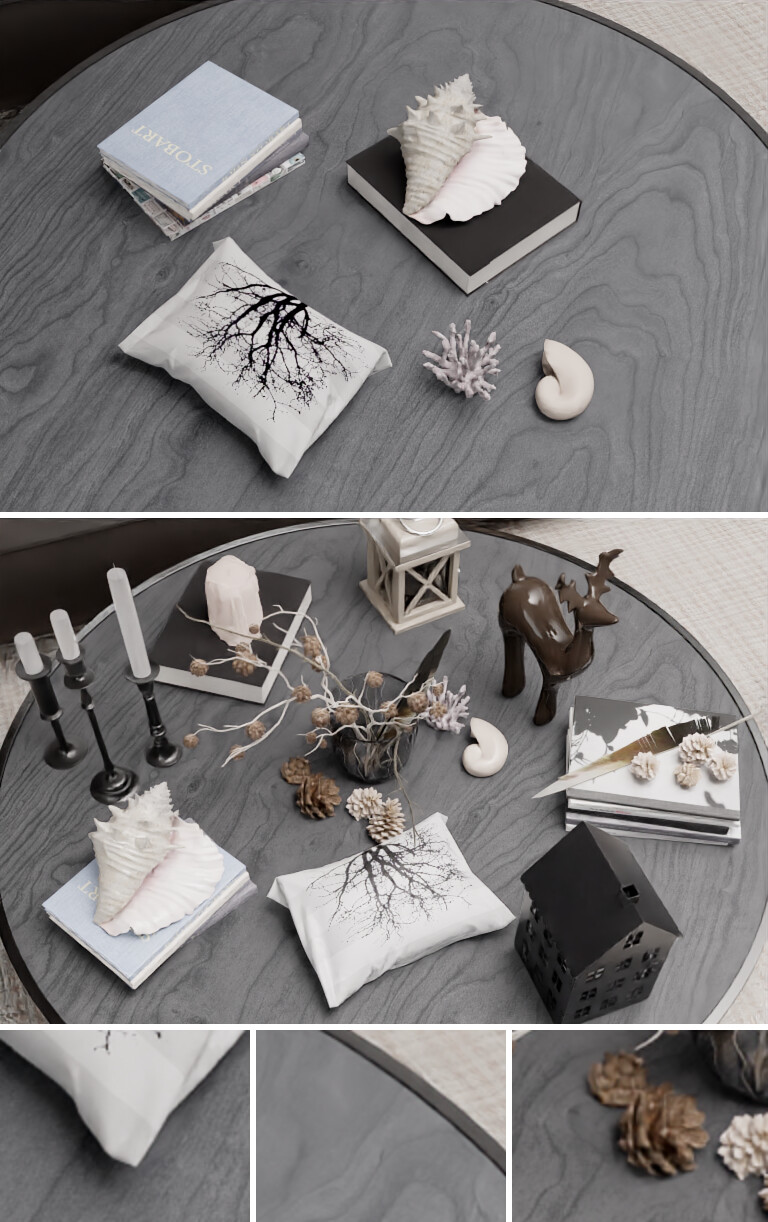} &
        \includegraphics[width=\mylength]{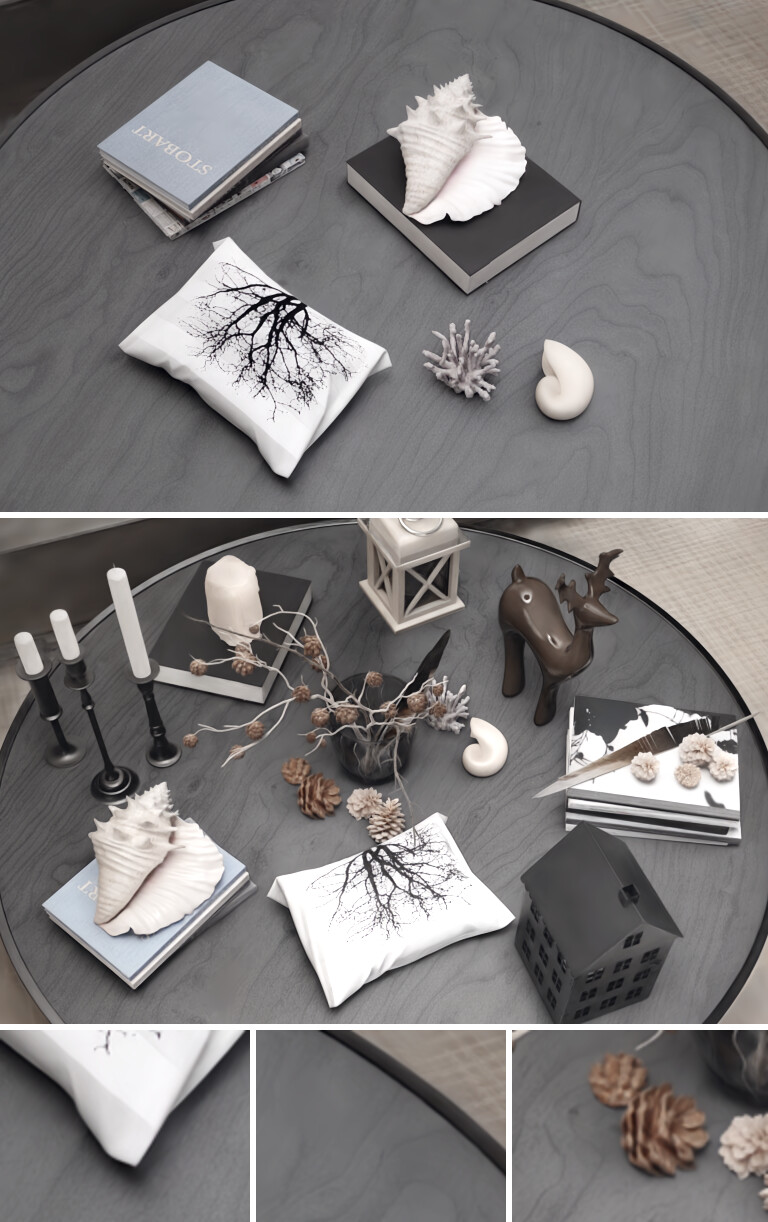} &
        \includegraphics[width=\mylength]{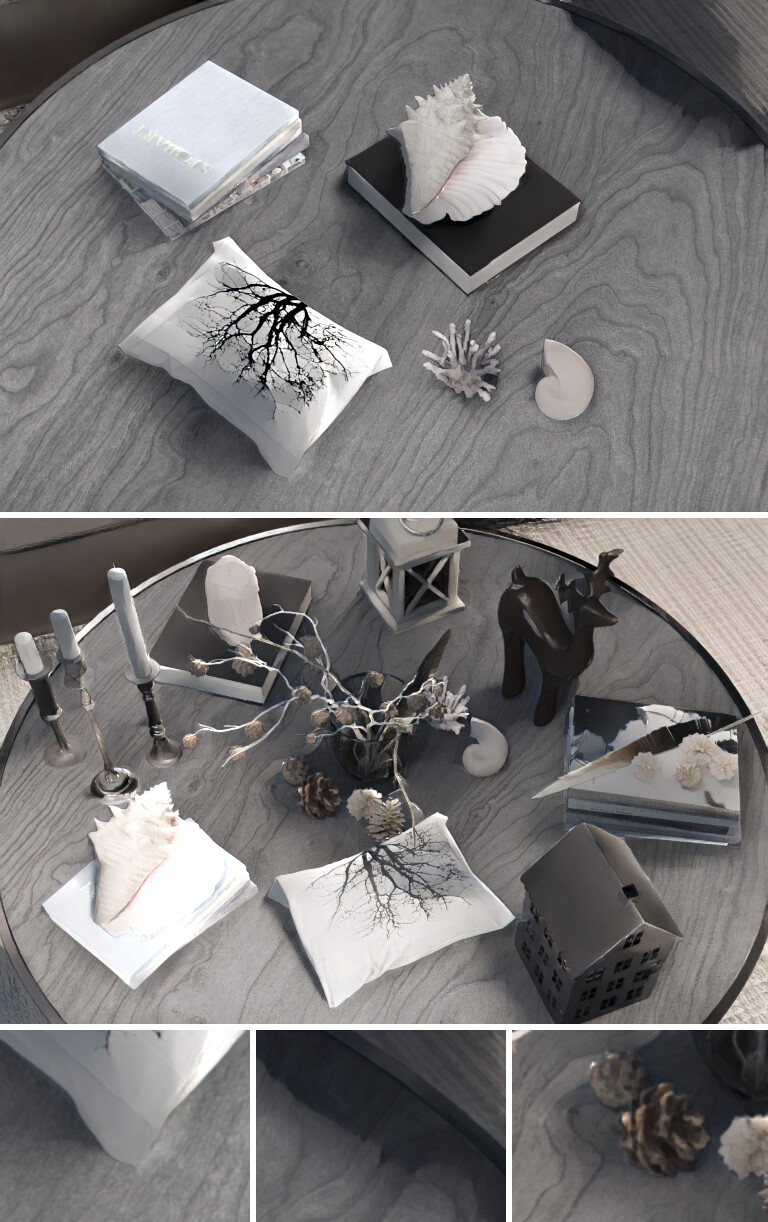} &
        \includegraphics[width=\mylength]{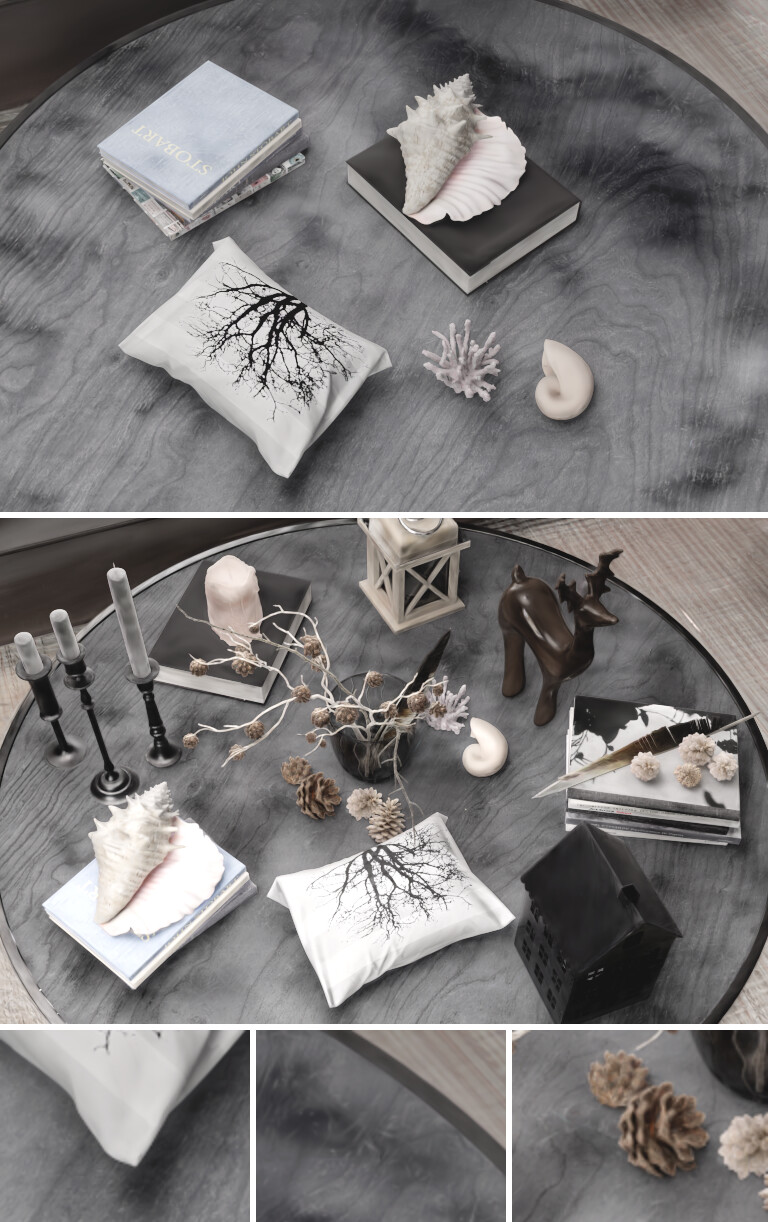} &
        \includegraphics[width=\mylength]{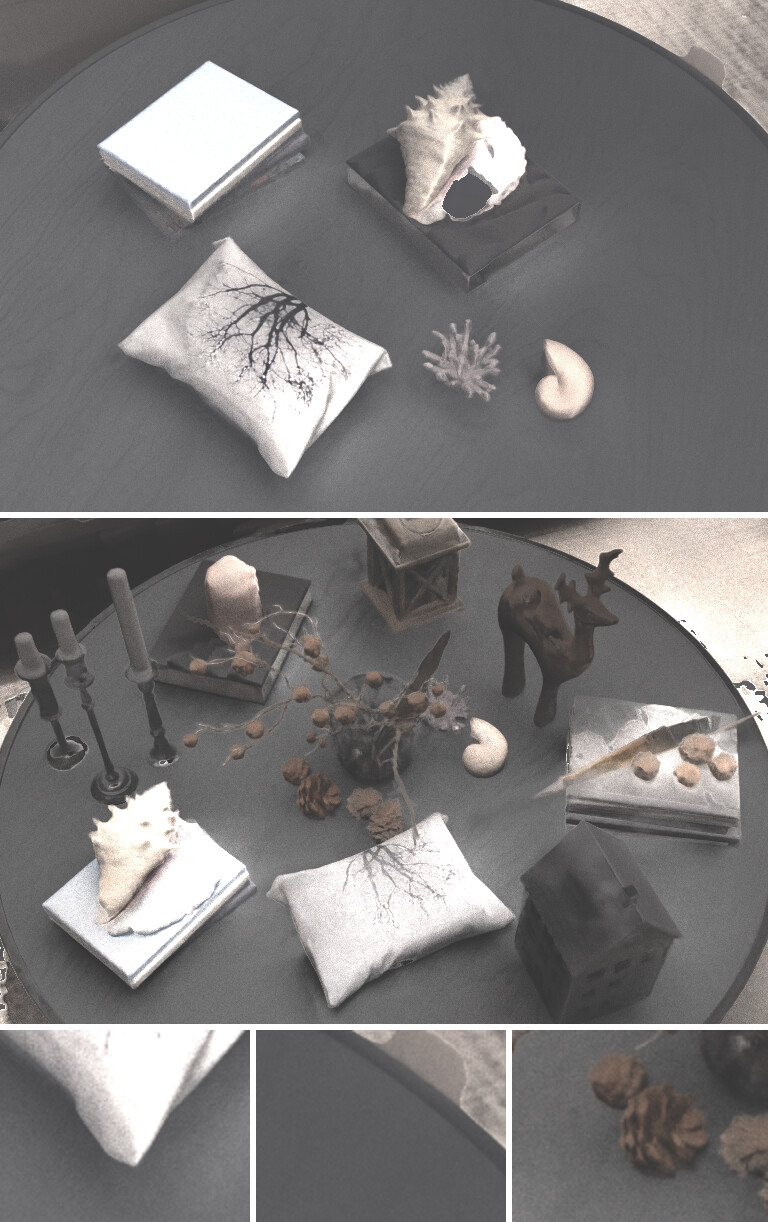} \\

    \end{tabular}
    \caption{
        \label{fig:synth_comp}
        We show comparative results of our method of synthetic scenes where the (approximate) ground truth is available (left), and compare to previous methods. Our approach is closer to the ground truth lighting, capturing the overall appearance in a realistic manner.
    }
    \end{figure*}

\begin{table*}[!h]
\setlength{\tabcolsep}{1.5pt}
\centering
\begin{tabular}{lccccccccccccc}
\toprule
Method $\rightarrow$ & \multicolumn{3}{c}{Ours}  & \multicolumn{3}{c}{Outcast~\cite{griffiths2022outcast}}  & \multicolumn{3}{c}{R3DGS~\cite{R3DG2023}}  & \multicolumn{3}{c}{TensoIR~\cite{jin2023tensoir}}  \\
Scene $\downarrow$ / Metrics & \footnotesize{PSNR}$_\uparrow$ & \footnotesize{LPIPS}$_\downarrow$ & \footnotesize{SSIM}$_\uparrow$ & \footnotesize{PSNR}$_\uparrow$ & \footnotesize{LPIPS}$_\downarrow$ & \footnotesize{SSIM}$_\uparrow$ & \footnotesize{PSNR}$_\uparrow$ & \footnotesize{LPIPS}$_\downarrow$ & \footnotesize{SSIM}$_\uparrow$ & \footnotesize{PSNR}$_\uparrow$ & \footnotesize{LPIPS}$_\downarrow$ & \footnotesize{SSIM}$_\uparrow$ \\
\midrule
Simple Bedroom & \cellcolor{red!25}20.57 & \cellcolor{red!25}0.156 & \cellcolor{red!25}0.868 & \cellcolor{yellow!25}17.24 & \cellcolor{yellow!25}0.207 & \cellcolor{yellow!25}0.808 & \cellcolor{orange!25}17.79 & \cellcolor{orange!25}0.174 & \cellcolor{orange!25}0.830 & 15.77 & 0.471 & 0.595 \\
Simple Kitchen & 17.45 & \cellcolor{red!25}0.154 & \cellcolor{red!25}0.855 & \cellcolor{yellow!25}17.91 & \cellcolor{yellow!25}0.205 & \cellcolor{orange!25}0.822 & \cellcolor{orange!25}18.55 & \cellcolor{orange!25}0.197 & \cellcolor{yellow!25}0.807 & \cellcolor{red!25}20.52 & 0.382 & 0.701 \\
Simple Livingroom & \cellcolor{red!25}22.12 & \cellcolor{orange!25}0.136 & \cellcolor{red!25}0.884 & \cellcolor{orange!25}21.09 & \cellcolor{red!25}0.125 & \cellcolor{orange!25}0.878 & \cellcolor{yellow!25}20.34 & \cellcolor{yellow!25}0.166 & \cellcolor{yellow!25}0.857 & 17.45 & 0.444 & 0.598 \\
Simple Office & \cellcolor{yellow!25}18.59 & \cellcolor{red!25}0.131 & \cellcolor{red!25}0.868 & \cellcolor{orange!25}18.97 & \cellcolor{yellow!25}0.196 & \cellcolor{orange!25}0.811 & \cellcolor{red!25}20.40 & \cellcolor{orange!25}0.173 & \cellcolor{yellow!25}0.808 & 18.22 & 0.446 & 0.644 \\
Complex Bedroom & \cellcolor{red!25}17.70 & \cellcolor{red!25}0.145 & \cellcolor{red!25}0.791 & \cellcolor{yellow!25}15.26 & \cellcolor{yellow!25}0.221 & \cellcolor{yellow!25}0.694 & \cellcolor{orange!25}16.69 & \cellcolor{orange!25}0.186 & \cellcolor{orange!25}0.741 & 14.42 & 0.434 & 0.555 \\
Complex Kitchen & \cellcolor{orange!25}19.28 & \cellcolor{red!25}0.152 & \cellcolor{red!25}0.811 & \cellcolor{yellow!25}18.44 & \cellcolor{yellow!25}0.178 & \cellcolor{orange!25}0.771 & \cellcolor{red!25}19.28 & \cellcolor{orange!25}0.168 & \cellcolor{yellow!25}0.755 & 16.70 & 0.471 & 0.533 \\
Complex Livingroom & \cellcolor{red!25}18.61 & \cellcolor{red!25}0.163 & \cellcolor{red!25}0.800 & \cellcolor{yellow!25}17.94 & \cellcolor{yellow!25}0.187 & \cellcolor{orange!25}0.783 & \cellcolor{orange!25}18.39 & \cellcolor{orange!25}0.175 & \cellcolor{yellow!25}0.770 & 16.82 & 0.382 & 0.602 \\
Complex Office & \cellcolor{red!25}20.20 & \cellcolor{red!25}0.096 & \cellcolor{red!25}0.858 & \cellcolor{yellow!25}17.22 & \cellcolor{yellow!25}0.169 & \cellcolor{orange!25}0.781 & \cellcolor{orange!25}18.93 & \cellcolor{orange!25}0.144 & \cellcolor{yellow!25}0.776 & 15.78 & 0.468 & 0.529 \\
\bottomrule
\end{tabular}

\caption{
\label{tab:3d-relighting}
Quantitative results of our 3D relighting on the synthetic datasets (where ground truth is available), compared to previous work, from left to right: OutCast~\cite{griffiths2022outcast} (run on individual images from 3DGS~\cite{gaussian-splats}), Relightable3DGaussians~\cite{R3DG2023}, and 
TensoIR~\cite{jin2023tensoir}. Arrows indicate higher/lower ($\uparrow/\downarrow$) is better. Results are color coded by \colorbox{red!25}{best}, \colorbox{orange!25}{second-} and \colorbox{yellow!25}{third-}best.}
\end{table*}



\paragraph*{Baselines.}
We compare our results to the method of \cite{philip2021free} which is specifically designed for complete scenes, TensoIR~\cite{jin2023tensoir} and RelightableGaussians~\cite{R3DG2023}. Given that most other methods do not handle full scenes well, we also create a new baseline, by first training 3DGS \cite{gaussian-splats} on the input data and render a test path using novel view synthesis; We then use OutCast~\cite{griffiths2022outcast} to relight each individual rendered frame using the target direction.
We trained TensoIR~\cite{jin2023tensoir} using the default configuration but modified the ``density\_shift'' parameter from $-10$ to $-8$ to achieve best results on our data. 
%
For Relightable 3D Gaussians~\cite{R3DG2023}, we train their ``Stage 1'' for 30K iterations and ``Stage 2'' for an additional 10K to recover the BRDF parameters. We then relight the scenes using 360° environment maps rendered in Blender using a generic empty room and a similar camera/flash setup for ground truth.
Finally, to improve the baselines we normalize the predictions of all methods; we first subtract the channel-wise mean and divide out the channel-wise standard deviation, and then multiply and add the corresponding parameters of the ground truths. These operations are performed in LAB space for all methods.

\paragraph*{Experimental methodology.}
We use our synthetic test scenes for providing quantitative results. To compare our method, we rendered 200 novel views with 18 different lighting directions to evaluate the relighting quality for each method by computing standard image quality metrics. Given the complexity of setup for \cite{philip2021free}, we only show qualitative results for 1 real scene in Fig.~\ref{fig:real-comp-qualitative}. Here, our method was trained at $768 \times 512$ resolution for 200k iterations, with a batch size of 8 and a learning rate of $10^{-4}$.

\paragraph*{Results.}
We present quantitative results in Table~\ref{tab:3d-relighting}. We present per-scene results on the following image quality metrics: PSNR, SSIM, and LPIPS~\cite{zhang2018perceptual}.
The results demonstrate that our method outperforms all others in all but a few scenarios, where it still achieves competitive performance.  


Qualitative comparisons are shown in Fig.~\ref{fig:synth_comp}; on the left we show the ground truth relit image rendered in Blender, and we then show our results, as well as those from Outcast~\cite{griffiths2022outcast}, Relightable 3D Gaussians~\cite{R3DG2023} and TensoIR~\cite{jin2023tensoir}. Please refer to the supplementary HTML viewer for more results. We clearly see that our method is closer to the ground truth, visually confirming the quantitative results in Tab.~\ref{tab:3d-relighting}.
TensoIR has difficulty reconstructing the geometry, and Relightable 3D Gaussians tend to have a ``splotchy'' look due to inaccurate normals. Outcast has difficulty with the overall lighting condition and can add incorrect shadows, but in many cases produces convincing results since it operates in image space. Our results show that by using the diffusion prior we manage to achieve realistic relighting, surpassing the state of the art.

\begin{figure*}[!h]
    \footnotesize
    \setlength{\tabcolsep}{0.24pt}
    \setlength{\mylength}{0.235\textwidth}
    \begin{tabular}{cccccc}
        \raisebox{98px}{\includegraphics[width=0.04\textwidth]{figures/gray_balls/left.png}} &
        \includegraphics[width=\mylength]{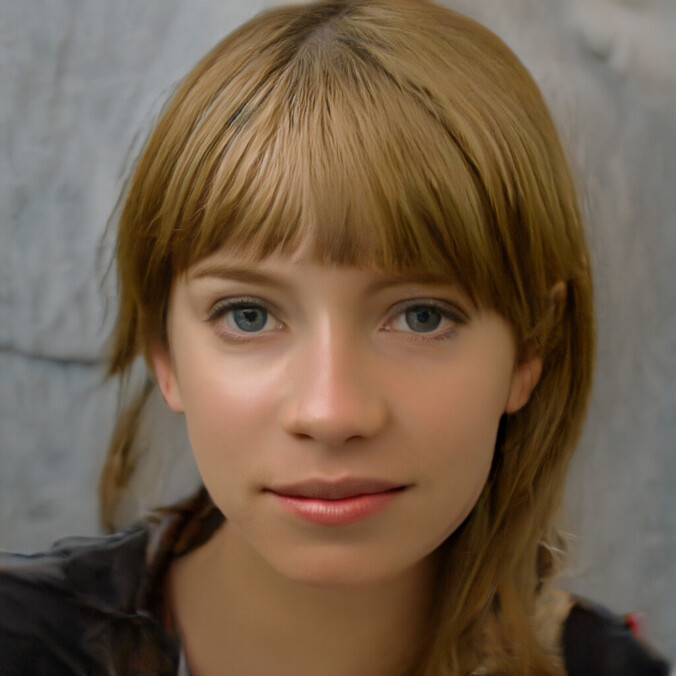} &
        \includegraphics[width=\mylength]{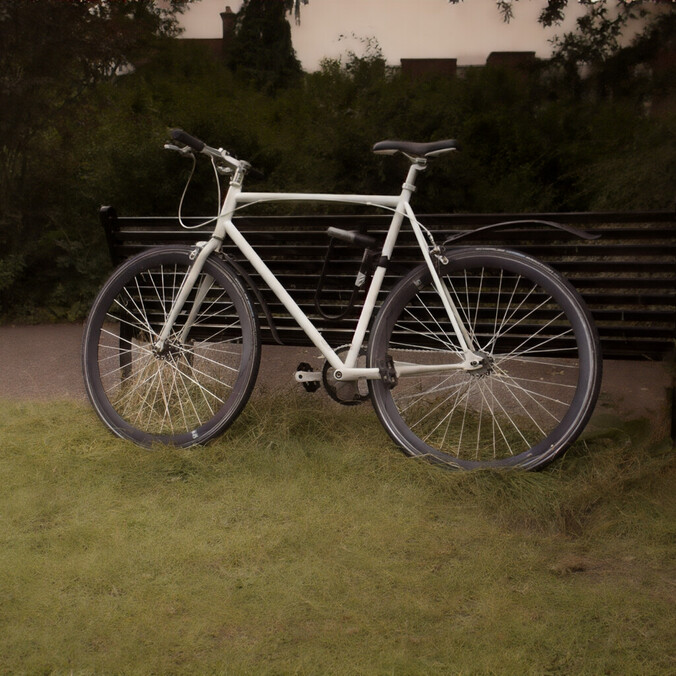} &
        \includegraphics[width=\mylength]{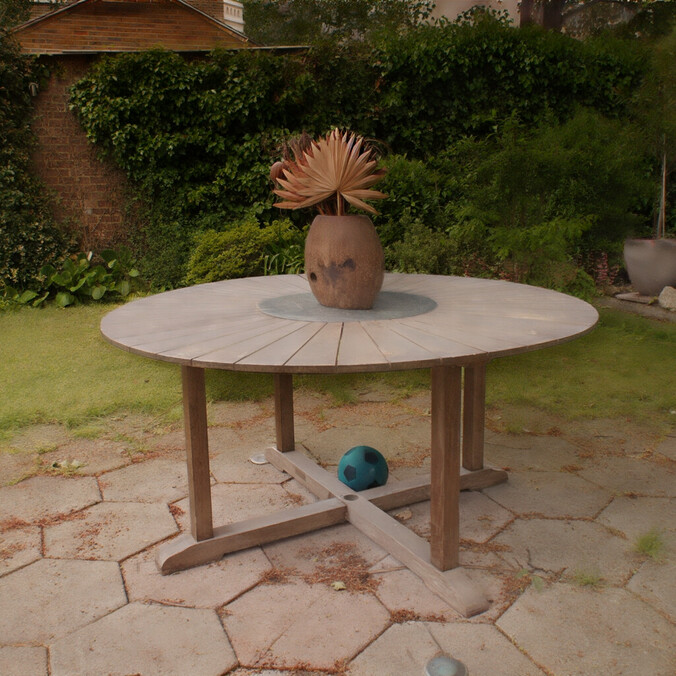} &
        \includegraphics[width=\mylength]{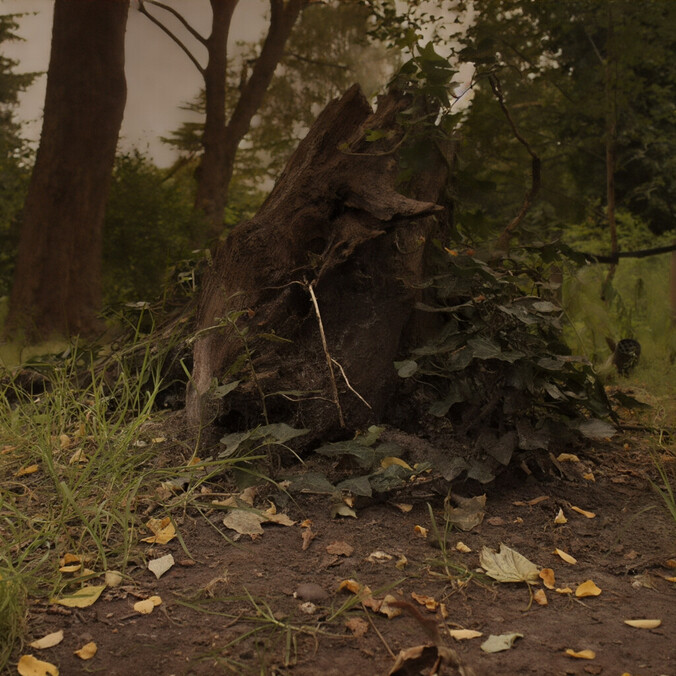} \\
        \raisebox{98px}{\includegraphics[width=0.04\textwidth]{figures/gray_balls/right.png}} &
        \includegraphics[width=\mylength]{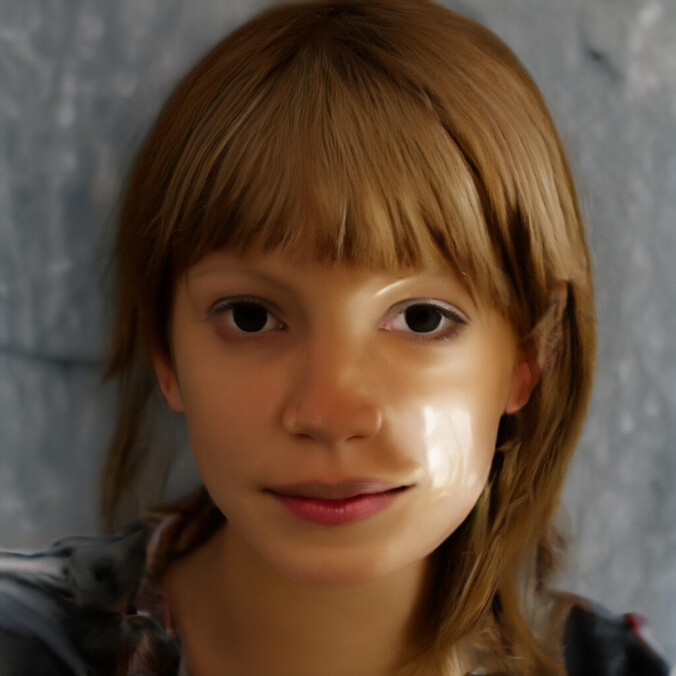} &
        \includegraphics[width=\mylength]{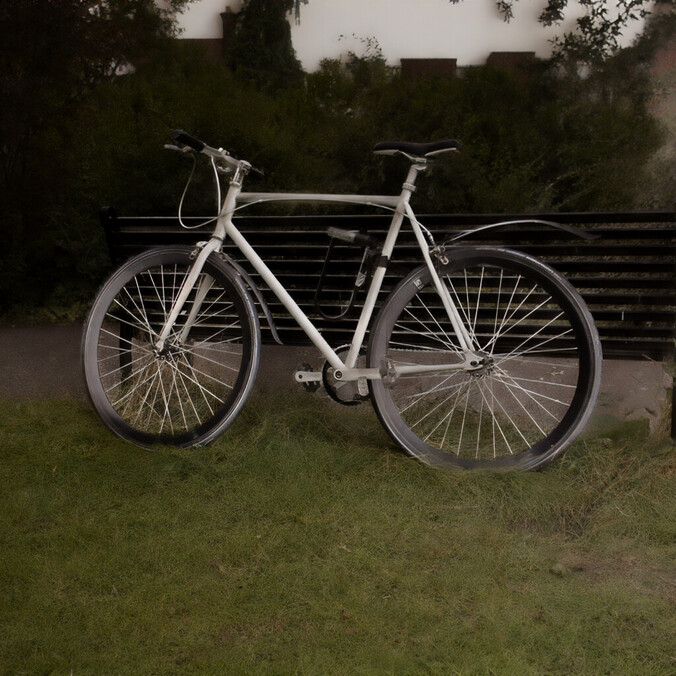} &
        \includegraphics[width=\mylength]{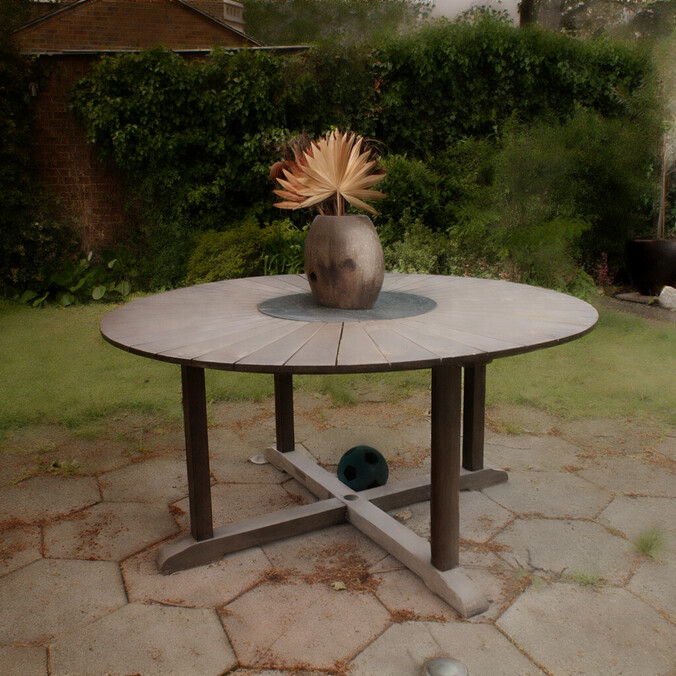} &
        \includegraphics[width=\mylength]{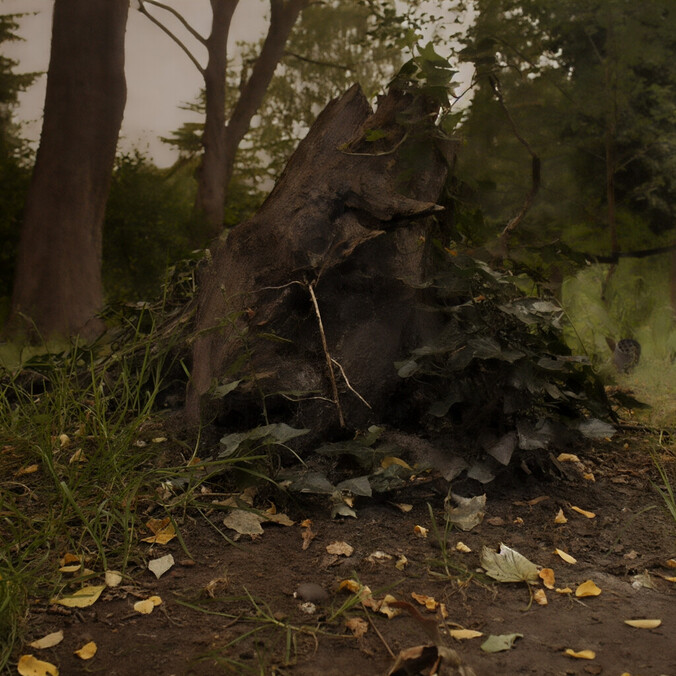} \\
        \raisebox{98px}{\includegraphics[width=0.04\textwidth]{figures/gray_balls/top.png}} &
        \includegraphics[width=\mylength]{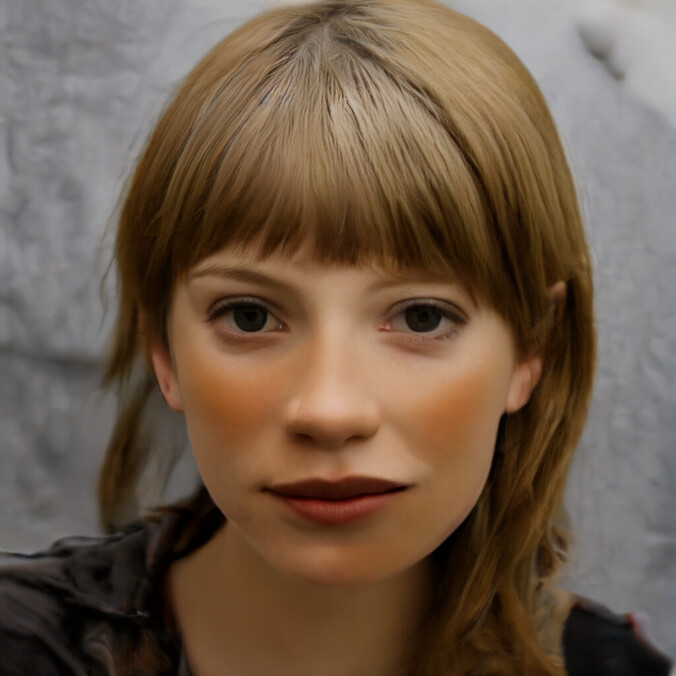} &
        \includegraphics[width=\mylength]{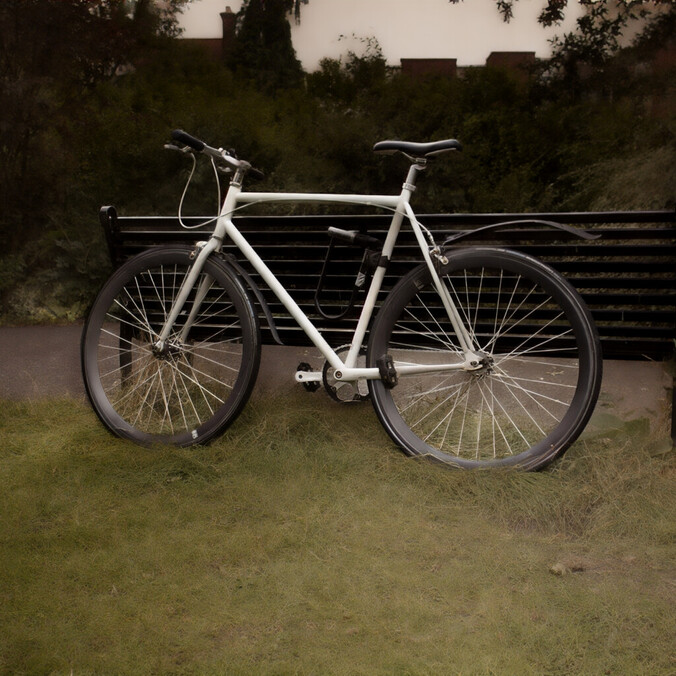} &
        \includegraphics[width=\mylength]{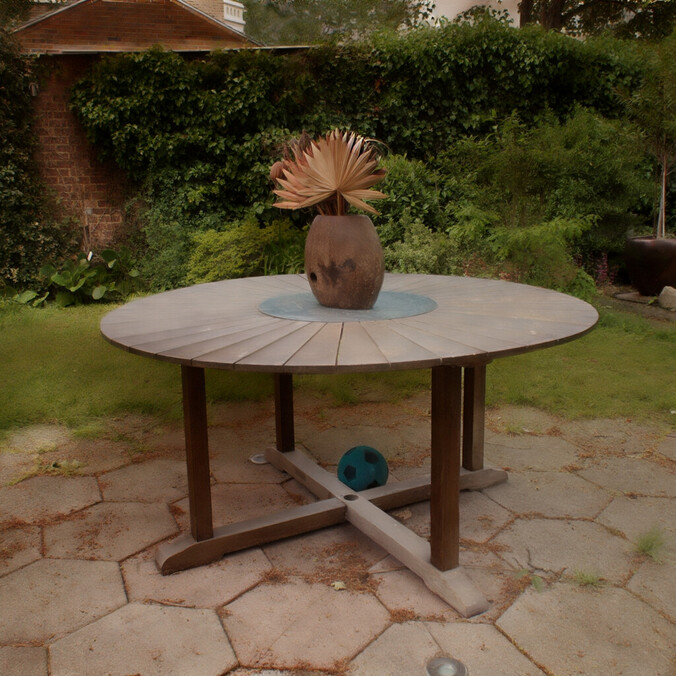} &
        \includegraphics[width=\mylength]{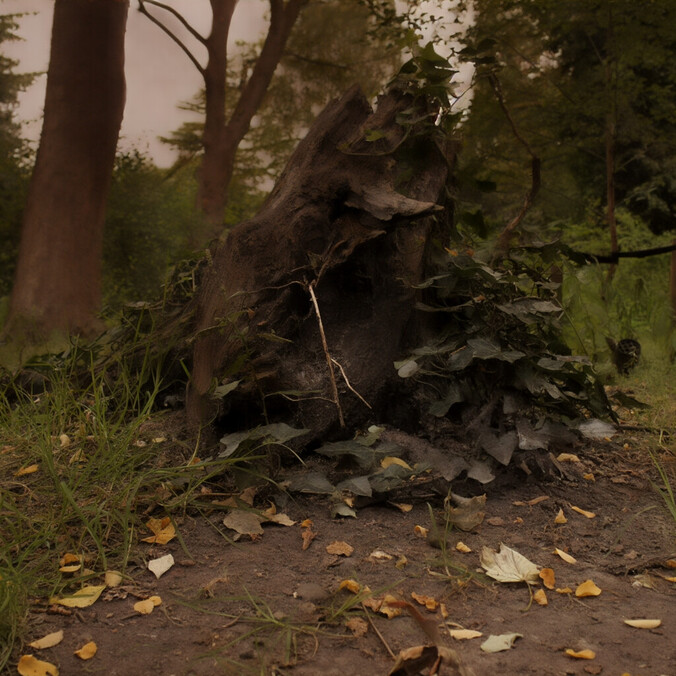} \\
        \raisebox{98px}{\includegraphics[width=0.04\textwidth]{figures/gray_balls/center.png}} &
        \includegraphics[width=\mylength]{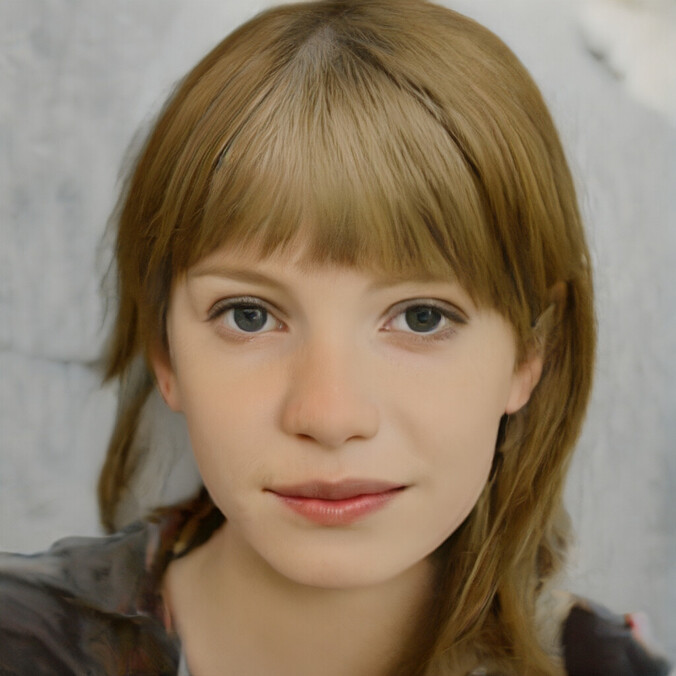} &
        \includegraphics[width=\mylength]{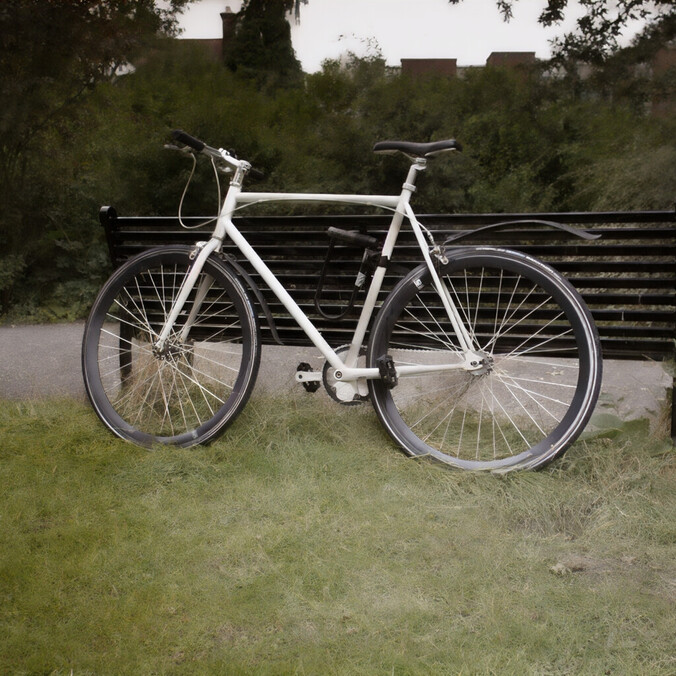} &
        \includegraphics[width=\mylength]{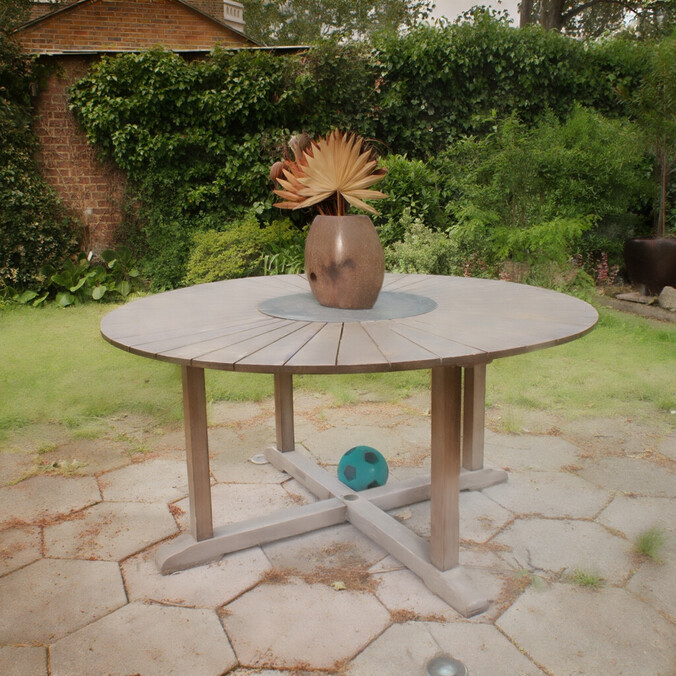} &
        \includegraphics[width=\mylength]{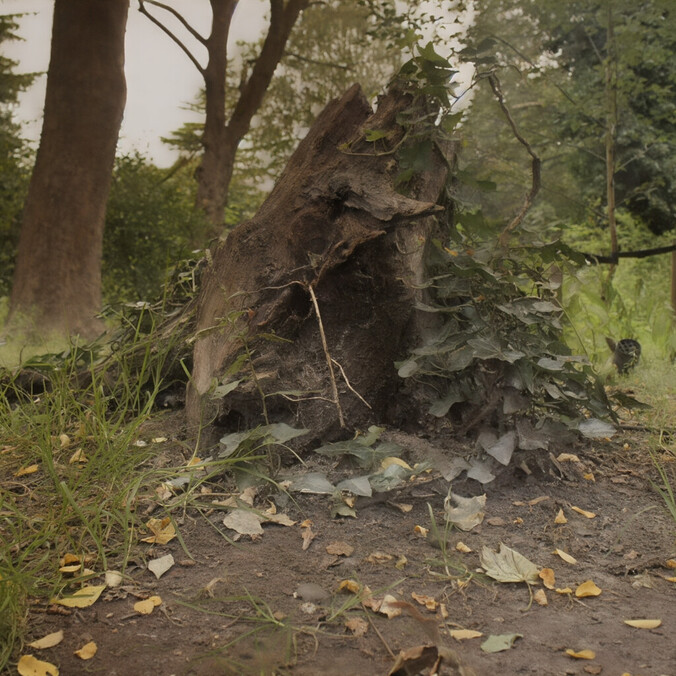} \\
    \end{tabular}
    \caption{
        \label{fig:ood_samples}
        \NEW{We show the results of our 2D relighting network on out-of-distribution images (StyleGAN-generated woman and MipNeRF360 \textsc{bicycle}, \textsc{garden}, and \textsc{stump}). On human faces, ControlNet may change the expression as well as the lighting, or create excessive shininess; on outdoor scenes, while the overall lighting direction is plausible, the network fails to generate sufficiently hard shadows.}
    }
\end{figure*}

\NEW{Our method was trained for indoor scenes; Fig.~\ref{fig:ood_samples} gives additional ControlNet results on out-of-distribution samples, showing that it can generalize to some extent to unseen scenes and lighting conditions, although the realism is lower than for in-distribution samples.}

\vspace{-1em}
\section{Conclusion}



We have presented the first method to effectively leverage the strong prior of large generative diffusion models in the context of radiance field relighting. 
Rather than relying on accurate geometry, material and/or lighting estimation, our approach models realistic illumination directly, by leveraging a general-purpose single-view, multi-illumination dataset and fine-tuning a large pretrained generative model.
Our results show that we can synthesize realistic relighting of captured scenes, while allowing interactive novel-view synthesis by building on such priors. 
Our method shows levels of realism for relighting that surpass the state of the art for cluttered indoor scenes (as opposed to isolated objects). 

\begin{figure}[h!]
    \includegraphics[width=\linewidth]{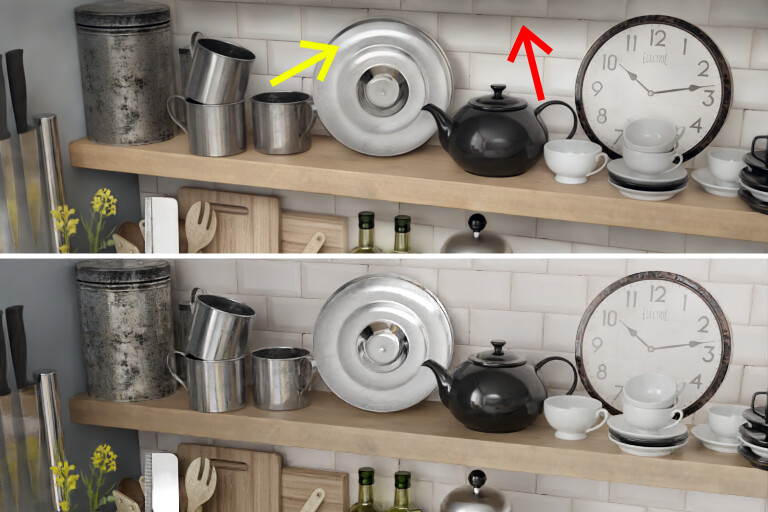}
    \caption{Example limitations of our approach, with our prediction (top) vs ground truth (bottom). Our ControlNet mistakenly produces a shadow at the top of the image while there should not be any \NEW{(red arrow)}, presumably assuming the presence of another top shelf. Additionally, the highlight position is somewhat incorrect \NEW{(yellow arrow)}, ostensibly because we define light direction in a manner that is not fully physically accurate.}
    \label{fig:limitations}
\end{figure}

One limitation of the proposed method is that it does not enforce physical accuracy: the target light direction is noisy and the ControlNet relies mostly on its powerful Stable Diffusion prior to relight rather than performing physics-based reasoning. For example, Fig.~\ref{fig:limitations} shows that ControlNet can hallucinate shadows due to unseen geometry, while there should not be any. Given that we define light direction in a manner that is not fully physically accurate, the positioning of highlight can be inaccurate, as is also shown in Fig.~\ref{fig:limitations}. In addition, the appearance embeddings can correct for global inconsistencies indirectly and do not explicitly rely on the learned 3D representation of the radiance field.
Our method does not always remove or move shadows in a fully accurate physically-based manner.
While our method clearly demonstrates that 2D diffusion model priors can be used for realistic relighting, the ability to perform more complex relighting---rather than just changing light direction---requires significant future research, e.g., by using more general training data as well as ways to encode and decode complex lighting. 

\NEW{An interesting direction for future work would be trying to enforce multi-view consistency more explicitly in ControlNet, e.g. by leveraging single-illumination multi-view data.} Another interesting direction is to develop solutions that would guide the predicted relighting making it more accurate, leveraging the 3D geometric information available in the radiance field more explicitly.


\paragraph*{Acknowledgements} This research was funded by the ERC Advanced grant FUNGRAPH No 788065 \url{http://fungraph.inria.fr/}, supported by NSERC grant DGPIN 2020-04799 and the Digital Research Alliance Canada. The authors are grateful to Adobe and NVIDIA for generous donations, and the OPAL infrastructure from Université Côte d’Azur. Thanks to Georgios Kopanas and Frédéric Fortier-Chouinard for helpful advice.


\bibliographystyle{eg-alpha}
\bibliography{bibliography}        


\end{document}